\pdfoutput=1

\documentclass[11pt]{article}

\usepackage{emnlp2021}

\usepackage{times}
\usepackage{latexsym}

\usepackage[T1]{fontenc}

\usepackage[utf8]{inputenc}

\usepackage{microtype}

\usepackage{amsmath}
\usepackage{amssymb}
\usepackage{graphicx}
\usepackage{transparent}
\usepackage{subcaption}
\usepackage{booktabs}
\usepackage{CJKutf8}
\usepackage[utf8]{inputenc}
\usepackage[main=english]{babel}
\usepackage{pdfcomment}
\usepackage{fancyvrb}
\VerbatimFootnotes

\newcommand{\boot}{\textsc{Boot}}

\newcommand{\spokenicon}{{\def\svgwidth{.65em}
\begingroup%
  \makeatletter%
  \providecommand\color[2][]{%
    \errmessage{(Inkscape) Color is used for the text in Inkscape, but the package 'color.sty' is not loaded}%
    \renewcommand\color[2][]{}%
  }%
  \providecommand\transparent[1]{%
    \errmessage{(Inkscape) Transparency is used (non-zero) for the text in Inkscape, but the package 'transparent.sty' is not loaded}%
    \renewcommand\transparent[1]{}%
  }%
  \providecommand\rotatebox[2]{#2}%
  \newcommand*\fsize{\dimexpr\f@size pt\relax}%
  \newcommand*\lineheight[1]{\fontsize{\fsize}{#1\fsize}\selectfont}%
  \ifx\svgwidth\undefined%
    \setlength{\unitlength}{384bp}%
    \ifx\svgscale\undefined%
      \relax%
    \else%
      \setlength{\unitlength}{\unitlength * \real{\svgscale}}%
    \fi%
  \else%
    \setlength{\unitlength}{\svgwidth}%
  \fi%
  \global\let\svgwidth\undefined%
  \global\let\svgscale\undefined%
  \makeatother%
  \begin{picture}(1,1)%
    \lineheight{1}%
    \setlength\tabcolsep{0pt}%
    \put(0,0){\includegraphics[width=\unitlength,page=1]{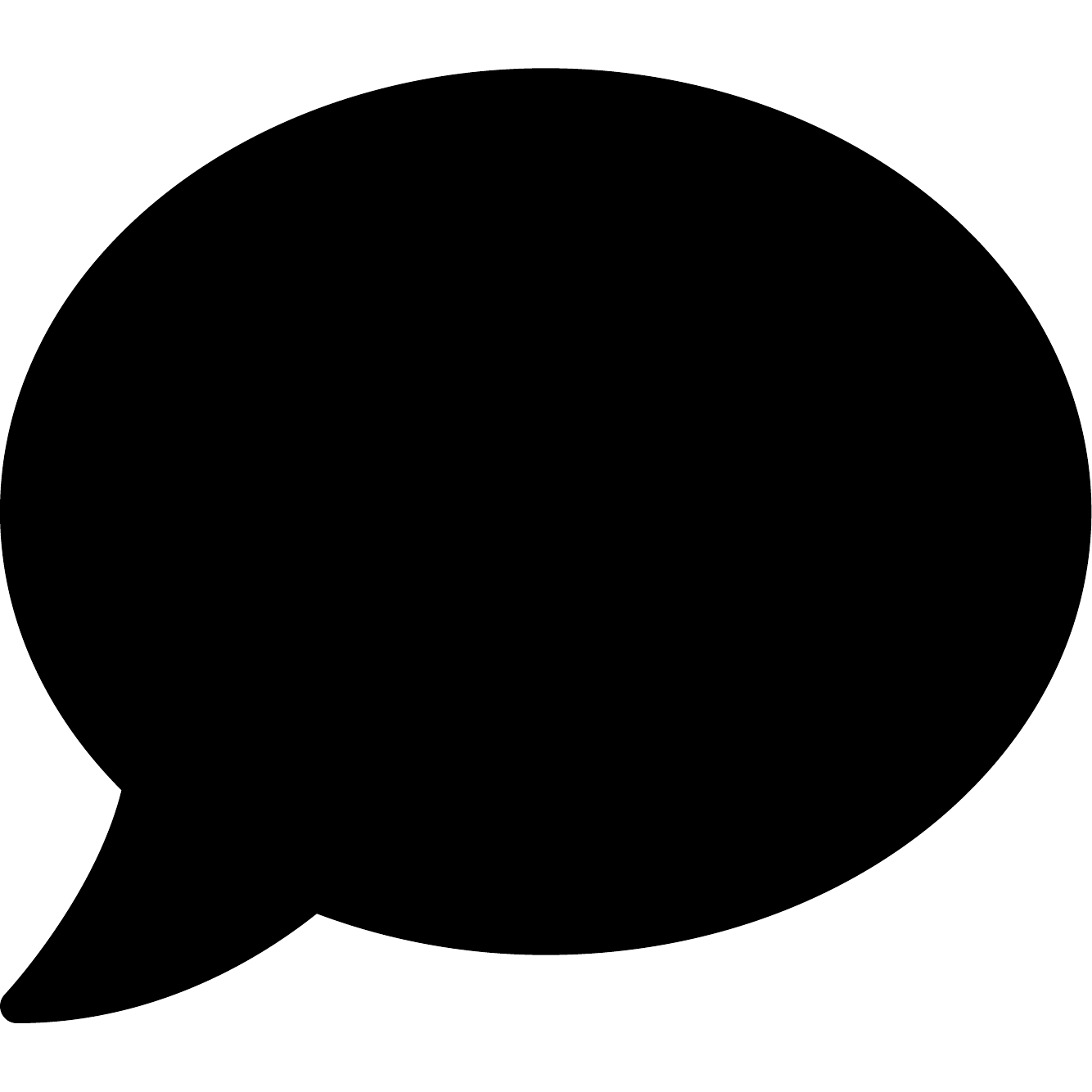}}%
  \end{picture}%
\endgroup%
}}
\newcommand{\fictionicon}{{\def\svgwidth{.6em}
\begingroup%
  \makeatletter%
  \providecommand\color[2][]{%
    \errmessage{(Inkscape) Color is used for the text in Inkscape, but the package 'color.sty' is not loaded}%
    \renewcommand\color[2][]{}%
  }%
  \providecommand\transparent[1]{%
    \errmessage{(Inkscape) Transparency is used (non-zero) for the text in Inkscape, but the package 'transparent.sty' is not loaded}%
    \renewcommand\transparent[1]{}%
  }%
  \providecommand\rotatebox[2]{#2}%
  \newcommand*\fsize{\dimexpr\f@size pt\relax}%
  \newcommand*\lineheight[1]{\fontsize{\fsize}{#1\fsize}\selectfont}%
  \ifx\svgwidth\undefined%
    \setlength{\unitlength}{336bp}%
    \ifx\svgscale\undefined%
      \relax%
    \else%
      \setlength{\unitlength}{\unitlength * \real{\svgscale}}%
    \fi%
  \else%
    \setlength{\unitlength}{\svgwidth}%
  \fi%
  \global\let\svgwidth\undefined%
  \global\let\svgscale\undefined%
  \makeatother%
  \begin{picture}(1,1.14285714)%
    \lineheight{1}%
    \setlength\tabcolsep{0pt}%
    \put(0,0){\includegraphics[width=\unitlength,page=1]{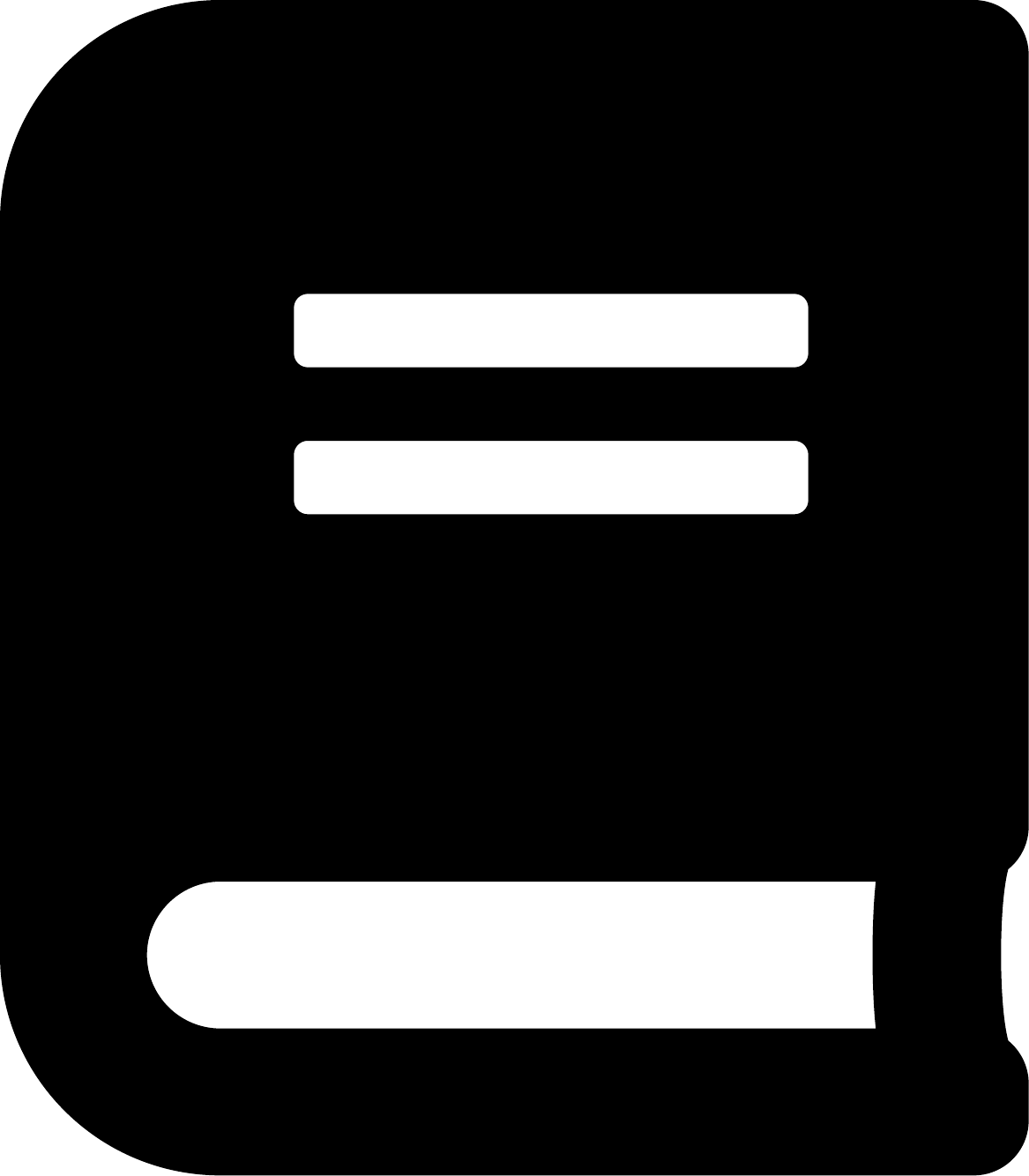}}%
  \end{picture}%
\endgroup%
}}
\newcommand{\newsicon}{{\def\svgwidth{.7em}
\begingroup%
  \makeatletter%
  \providecommand\color[2][]{%
    \errmessage{(Inkscape) Color is used for the text in Inkscape, but the package 'color.sty' is not loaded}%
    \renewcommand\color[2][]{}%
  }%
  \providecommand\transparent[1]{%
    \errmessage{(Inkscape) Transparency is used (non-zero) for the text in Inkscape, but the package 'transparent.sty' is not loaded}%
    \renewcommand\transparent[1]{}%
  }%
  \providecommand\rotatebox[2]{#2}%
  \newcommand*\fsize{\dimexpr\f@size pt\relax}%
  \newcommand*\lineheight[1]{\fontsize{\fsize}{#1\fsize}\selectfont}%
  \ifx\svgwidth\undefined%
    \setlength{\unitlength}{432bp}%
    \ifx\svgscale\undefined%
      \relax%
    \else%
      \setlength{\unitlength}{\unitlength * \real{\svgscale}}%
    \fi%
  \else%
    \setlength{\unitlength}{\svgwidth}%
  \fi%
  \global\let\svgwidth\undefined%
  \global\let\svgscale\undefined%
  \makeatother%
  \begin{picture}(1,0.88888889)%
    \lineheight{1}%
    \setlength\tabcolsep{0pt}%
    \put(0,0){\includegraphics[width=\unitlength,page=1]{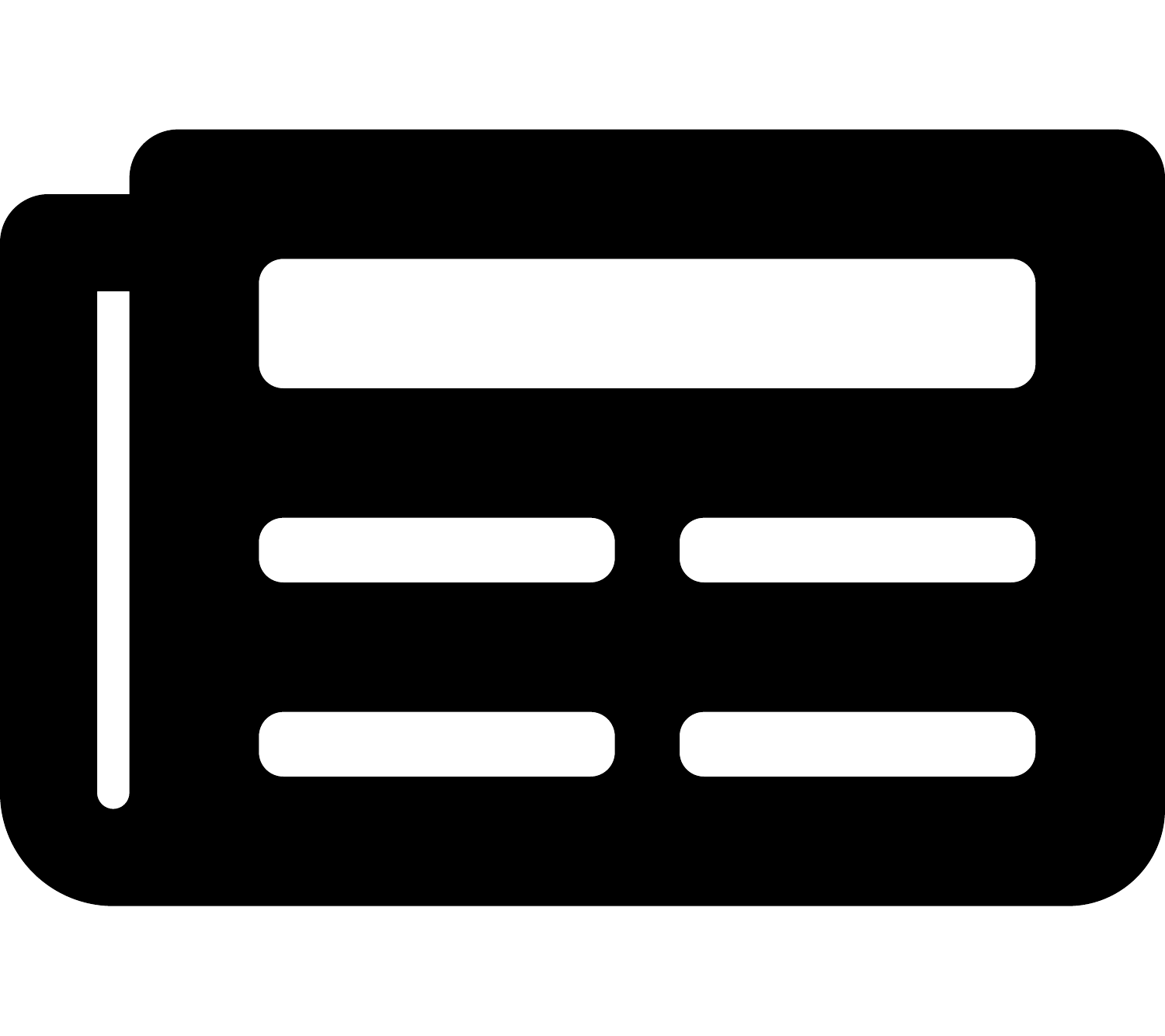}}%
  \end{picture}%
\endgroup%
}}
\newcommand{\wikiicon}{{\def\svgwidth{.7em}
\begingroup%
  \makeatletter%
  \providecommand\color[2][]{%
    \errmessage{(Inkscape) Color is used for the text in Inkscape, but the package 'color.sty' is not loaded}%
    \renewcommand\color[2][]{}%
  }%
  \providecommand\transparent[1]{%
    \errmessage{(Inkscape) Transparency is used (non-zero) for the text in Inkscape, but the package 'transparent.sty' is not loaded}%
    \renewcommand\transparent[1]{}%
  }%
  \providecommand\rotatebox[2]{#2}%
  \newcommand*\fsize{\dimexpr\f@size pt\relax}%
  \newcommand*\lineheight[1]{\fontsize{\fsize}{#1\fsize}\selectfont}%
  \ifx\svgwidth\undefined%
    \setlength{\unitlength}{480bp}%
    \ifx\svgscale\undefined%
      \relax%
    \else%
      \setlength{\unitlength}{\unitlength * \real{\svgscale}}%
    \fi%
  \else%
    \setlength{\unitlength}{\svgwidth}%
  \fi%
  \global\let\svgwidth\undefined%
  \global\let\svgscale\undefined%
  \makeatother%
  \begin{picture}(1,0.8)%
    \lineheight{1}%
    \setlength\tabcolsep{0pt}%
    \put(0,0){\includegraphics[width=\unitlength,page=1]{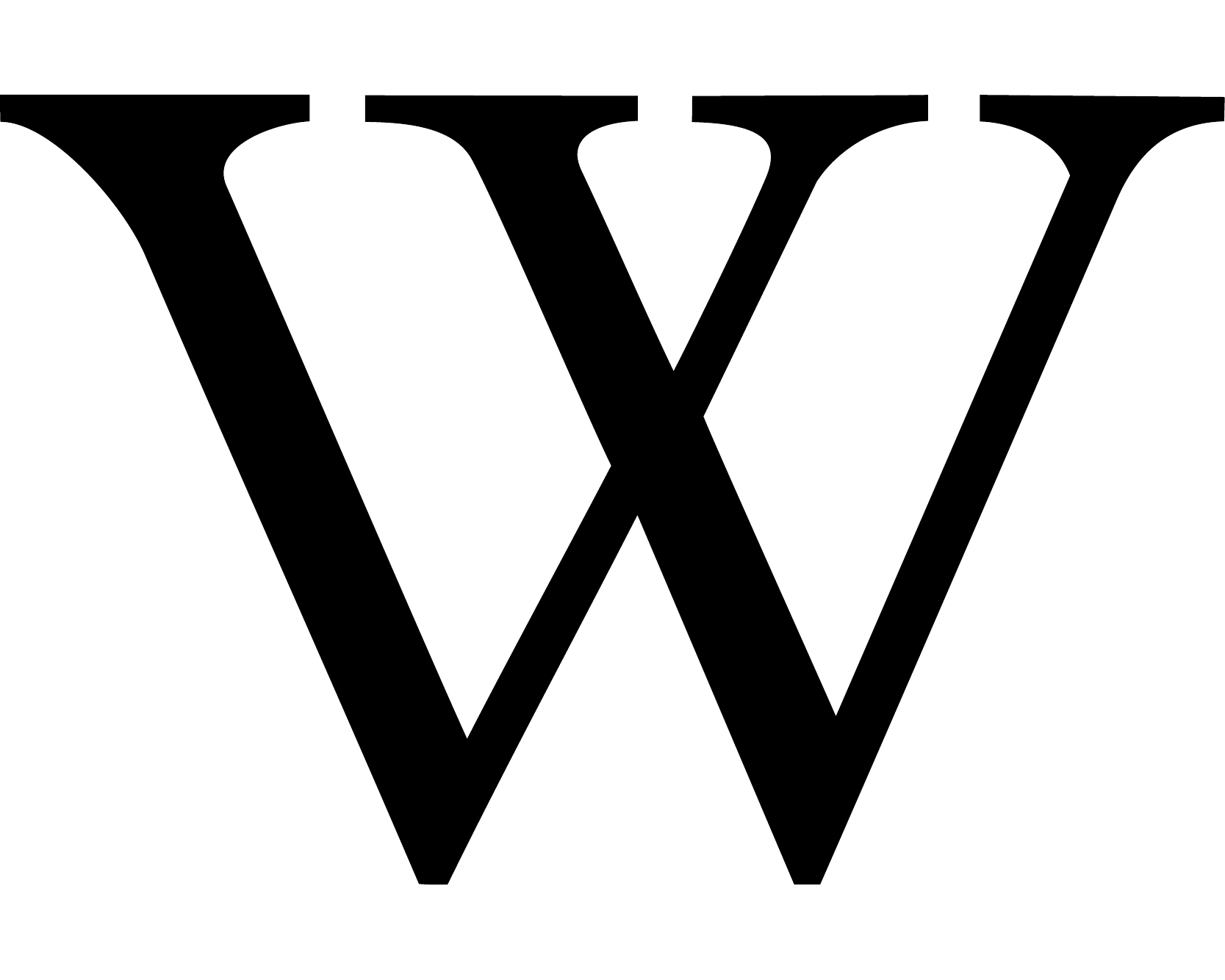}}%
  \end{picture}%
\endgroup%
}}
\newcommand{\grammaricon}{{\def\svgwidth{.6em}
\begingroup%
  \makeatletter%
  \providecommand\color[2][]{%
    \errmessage{(Inkscape) Color is used for the text in Inkscape, but the package 'color.sty' is not loaded}%
    \renewcommand\color[2][]{}%
  }%
  \providecommand\transparent[1]{%
    \errmessage{(Inkscape) Transparency is used (non-zero) for the text in Inkscape, but the package 'transparent.sty' is not loaded}%
    \renewcommand\transparent[1]{}%
  }%
  \providecommand\rotatebox[2]{#2}%
  \newcommand*\fsize{\dimexpr\f@size pt\relax}%
  \newcommand*\lineheight[1]{\fontsize{\fsize}{#1\fsize}\selectfont}%
  \ifx\svgwidth\undefined%
    \setlength{\unitlength}{384bp}%
    \ifx\svgscale\undefined%
      \relax%
    \else%
      \setlength{\unitlength}{\unitlength * \real{\svgscale}}%
    \fi%
  \else%
    \setlength{\unitlength}{\svgwidth}%
  \fi%
  \global\let\svgwidth\undefined%
  \global\let\svgscale\undefined%
  \makeatother%
  \begin{picture}(1,1)%
    \lineheight{1}%
    \setlength\tabcolsep{0pt}%
    \put(0,0){\includegraphics[width=\unitlength,page=1]{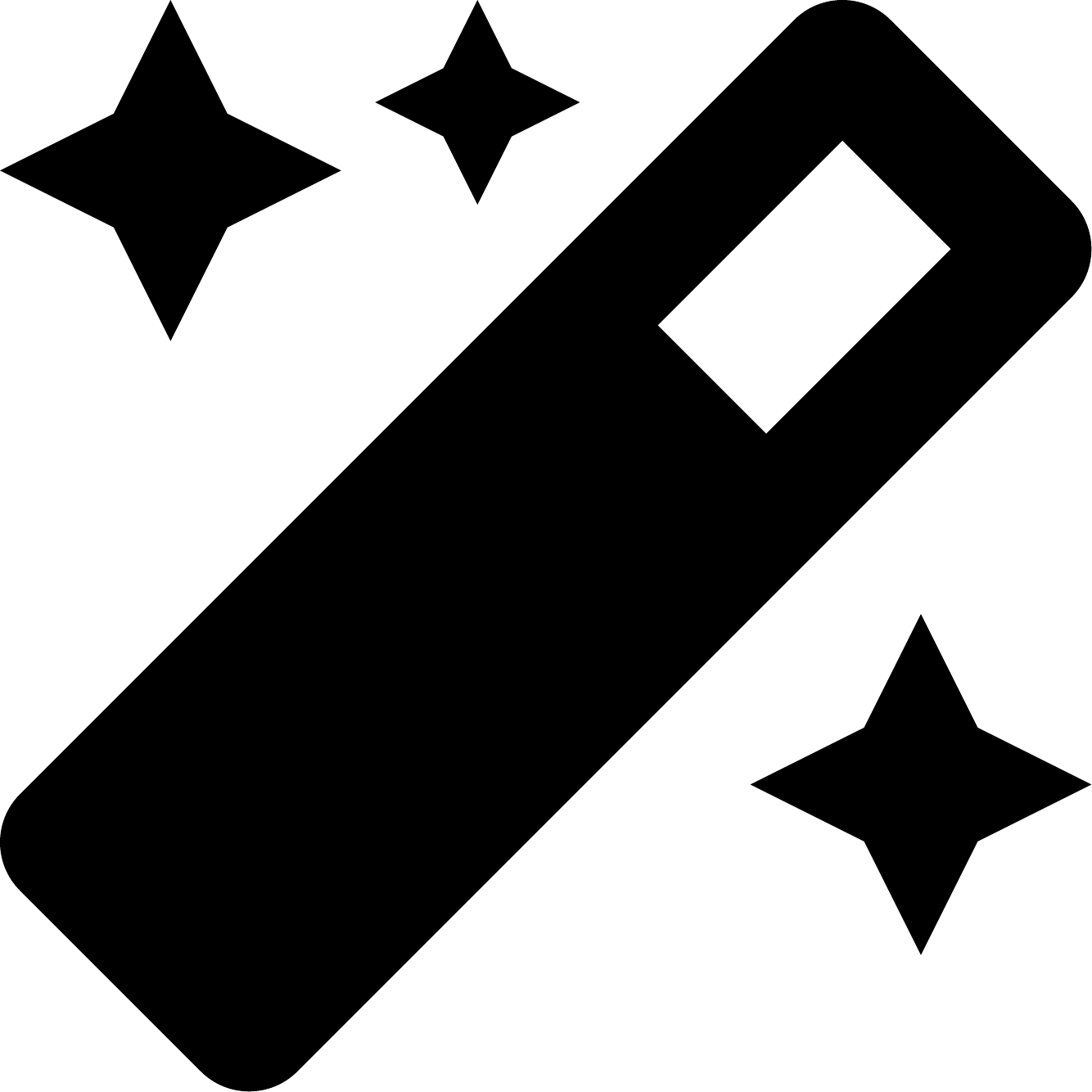}}%
  \end{picture}%
\endgroup%
}}
\newcommand{\socialicon}{{\def\svgwidth{.6em}
\begingroup%
  \makeatletter%
  \providecommand\color[2][]{%
    \errmessage{(Inkscape) Color is used for the text in Inkscape, but the package 'color.sty' is not loaded}%
    \renewcommand\color[2][]{}%
  }%
  \providecommand\transparent[1]{%
    \errmessage{(Inkscape) Transparency is used (non-zero) for the text in Inkscape, but the package 'transparent.sty' is not loaded}%
    \renewcommand\transparent[1]{}%
  }%
  \providecommand\rotatebox[2]{#2}%
  \newcommand*\fsize{\dimexpr\f@size pt\relax}%
  \newcommand*\lineheight[1]{\fontsize{\fsize}{#1\fsize}\selectfont}%
  \ifx\svgwidth\undefined%
    \setlength{\unitlength}{336bp}%
    \ifx\svgscale\undefined%
      \relax%
    \else%
      \setlength{\unitlength}{\unitlength * \real{\svgscale}}%
    \fi%
  \else%
    \setlength{\unitlength}{\svgwidth}%
  \fi%
  \global\let\svgwidth\undefined%
  \global\let\svgscale\undefined%
  \makeatother%
  \begin{picture}(1,1.14285714)%
    \lineheight{1}%
    \setlength\tabcolsep{0pt}%
    \put(0,0){\includegraphics[width=\unitlength,page=1]{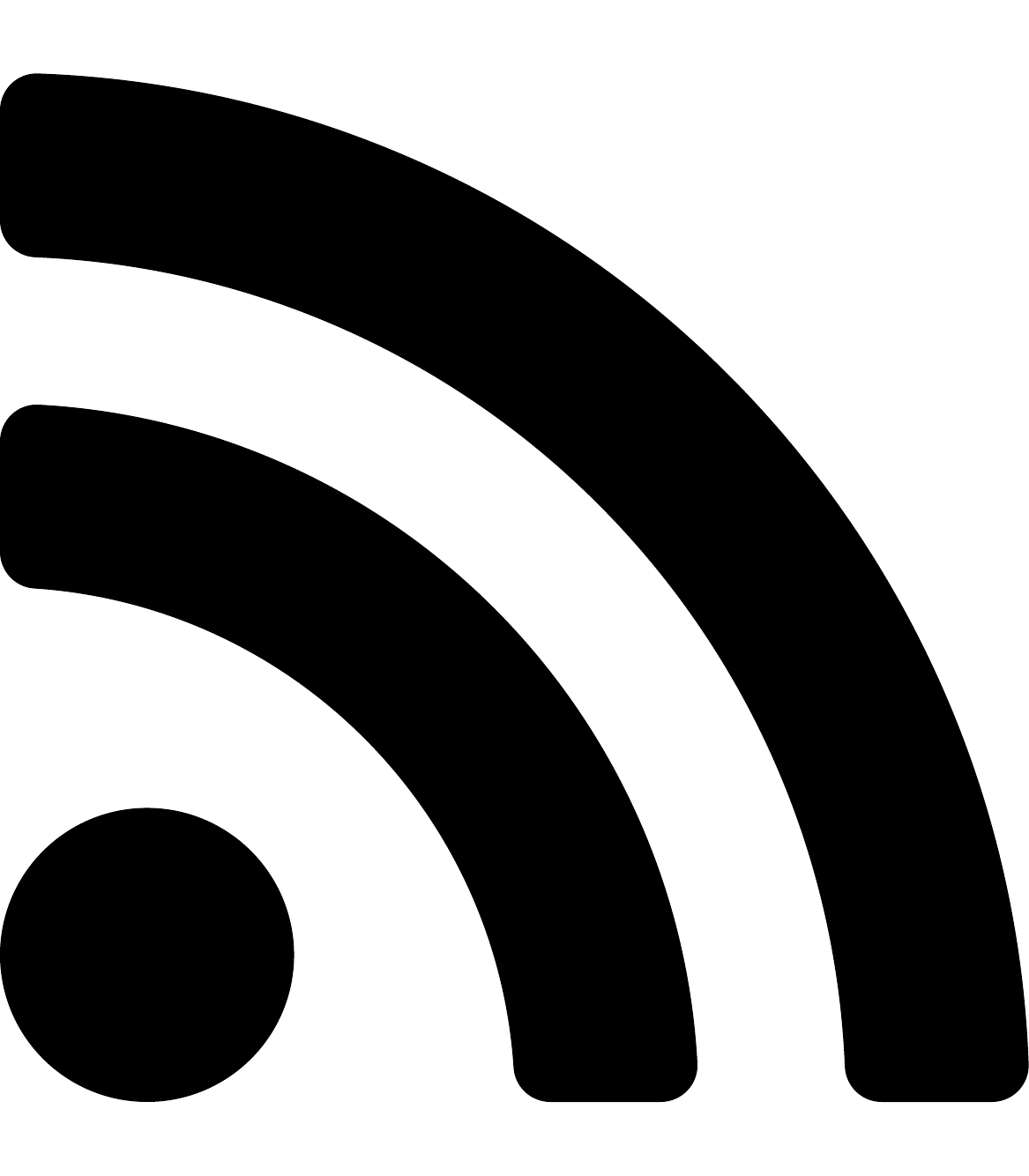}}%
  \end{picture}%
\endgroup%
}}

\title{Genre as Weak Supervision for Cross-lingual Dependency Parsing}

\author{Max M{\"u}ller-Eberstein \and Rob van der Goot \and Barbara Plank \\
  Department of Computer Science \\
  IT University of Copenhagen, Denmark \\
  \texttt{mamy@itu.dk, robv@itu.dk, bapl@itu.dk}}

\begin{document}

\maketitle

\begin{abstract}
Recent work has shown that monolingual masked language models learn to represent data-driven notions of language variation which can be used for domain-targeted training data selection.
Dataset \textit{genre} labels are already frequently available, yet remain largely unexplored in cross-lingual setups. We harness this genre metadata as a weak supervision signal for targeted data selection in zero-shot dependency parsing. Specifically, we project treebank-level \textit{genre} information to the finer-grained sentence level, with the goal to amplify information implicitly stored in unsupervised contextualized representations.
We demonstrate that genre is recoverable from multilingual contextual embeddings and that it provides an effective signal for training data selection in cross-lingual, zero-shot scenarios. For 12 low-resource language treebanks, six of which are test-only, our genre-specific methods significantly outperform competitive baselines as well as recent embedding-based methods for data selection. Moreover, genre-based data selection provides new state-of-the-art results for three of these target languages.
\end{abstract}

\section{Introduction}

Multilingual masked language models (MLMs) trained on immense quantities of heterogeneous texts \citep{devlin2019, brown2020,conneau-etal-2020-unsupervised} have recently made applications such as highly cross-lingual dependency parsing a reality (\citealp{kondratyuk-straka-2019-75}). Adjacently, it has also been recognized that they capture characteristics relevant for training data selection \citep{aharoni2020} and can be efficiently fine-tuned for higher task-specific performance \citep{gururangan2020, dai2020, lauscher2020, ustun-etal-2020-udapter}. These considerations are especially important in computationally restricted environments and when data from the target distribution are unavailable.

Universal Dependencies (\citealp{nivre-etal-2020-universal}; UD) provides an extensive testing ground for such scenarios: Its language diversity is constantly increasing (from 10 in v1.0 to 104 in v2.7) and low-resource languages are often limited to a single test-set-only treebank. As most of the 7,000+ languages in the world similarly lack any annotated training data, effective zero-shot transfer learning is crucial for achieving wider linguistic coverage.

\begin{figure}
    \centering
    \vspace{-1em}
    \pdftooltip{\includegraphics[width=.48\textwidth]{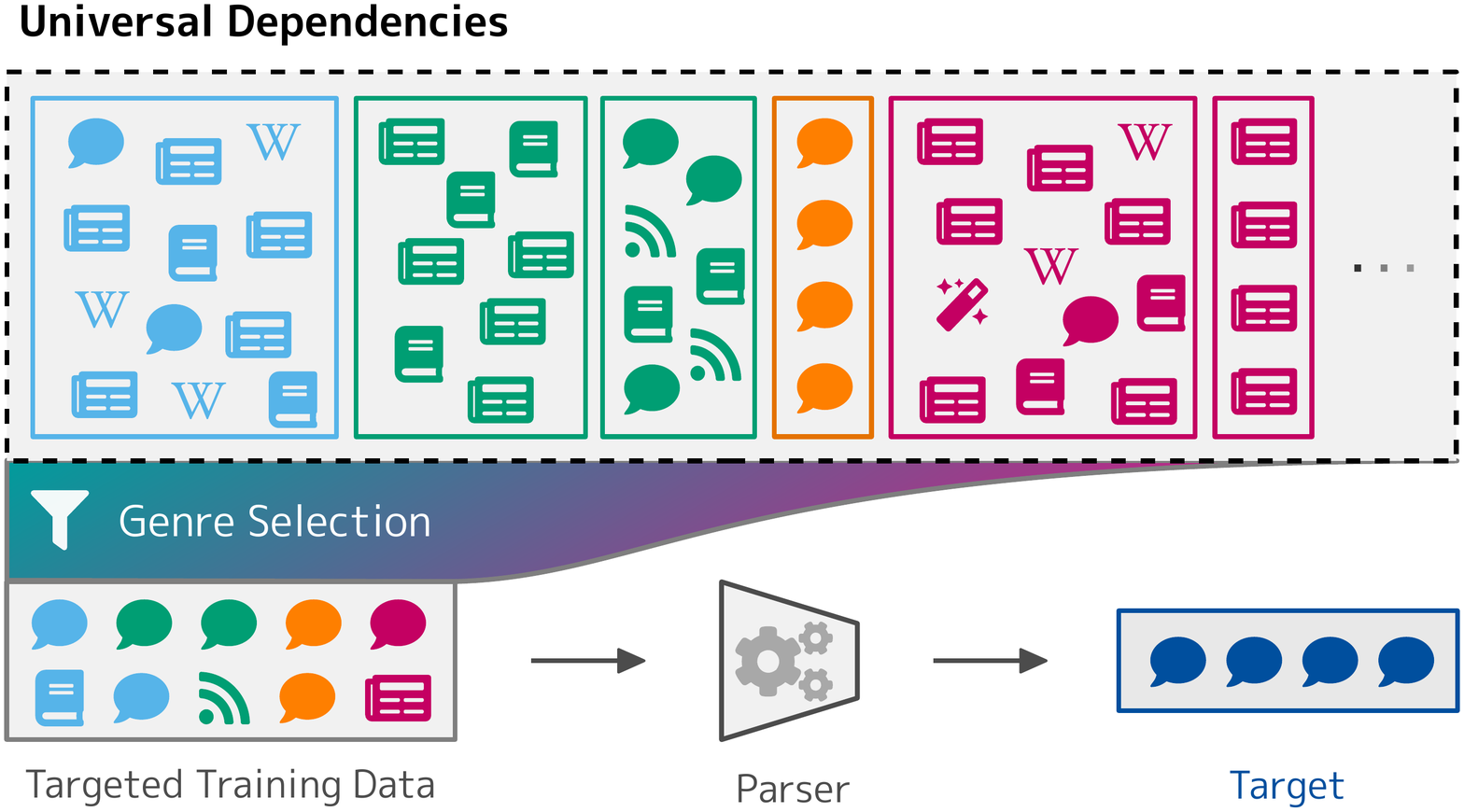}}{Screenreader Caption: Universal Dependencies contain treebanks in multiple languages containing instances from different genres (some treebanks contain instances from only one genre, but the majority contain multiple). In order to train a parser for a target treebank without in-language data in UD, genre-driven selection creates a targeted, out-of-language training corpus containing mostly instances in the target genre (e.g. spoken).}
    \vspace{-2.5em}
    \caption{\textbf{Genre-driven Training Data Selection} for a zero-shot target treebank. In absence of annotated in-language data, we propose \textit{genre} as a weak supervision signal for targeted instance selection from a large pool of out-of-language treebanks.}
    \label{fig:method-overview}
\end{figure}

Criteria for selecting training data within such settings vary, and a practitioner may determine relevance by proxy of language relatedness or treebank content. This leads us to the question: If our goal is to develop a parser for a known domain in an unseen language, can a signal such as \textit{genre} guide our selection of cross-lingual training data from a significantly larger, diverse pool (Figure \ref{fig:method-overview})?

Within the heterogeneity of written and spoken (transcribed) data, genre broadly encompasses variation along the functional role of a text~\cite{kessler-etal-1997-automatic}. A clear definition is complex if not impossible and communities refer to genre, domain, style or register in different ways~\cite{kessler-etal-1997-automatic,lee2001genres,webber-2009-genre,plank2011domain}. In this work, we take a pragmatic approach and use genre as defined by the 18 community-provided categories in UD (\citealp{ud27}). 
These genres are assigned at the treebank level and ``are neither mutually exclusive nor based on homogeneous criteria, but [are] currently the best documentation that can be obtained'' \citep{nivre-etal-2020-universal}.

\paragraph{Contributions} In order to facilitate finer-grained, instance-level data selection for cross-lingual parsing in absence of in-language training data, we provide three contributions:

First, we provide an analysis of the genre distribution in UD v2.7 \citep{ud27} across 104 languages and 177 treebanks (Section \ref{sec:ud-genre}).

Next, we introduce three targeted data selection strategies which amplify existing genre information in multilingual contextualized embeddings in order to enable sentence-level selection based on UD's treebank-level genre annotations (Section \ref{sec:method}).

Finally, we apply the extracted genre information to proxy training data selection for 12 typologically diverse low-resource treebanks. In absence of any in-language training data, our approach outperforms selection using treebank metadata alone as well as purely embedding-based instance selection and surpasses state-of-the-art results on three treebanks (Section \ref{sec:experiments}).\footnote{Code at \href{https://personads.me/x/emnlp-2021-code}{https://personads.me/x/emnlp-2021-code}.}

\section{Related Work}\label{sec:relwork}

Despite advances in zero-shot performance \citep{devlin2019, brown2020} and increasingly cross-lingual parsers~\citep{kondratyuk-straka-2019-75}, fine-tuning has remained a crucial step for achieving state-of-the-art performance. \citet{meechan2019} demonstrate that this holds true for low-resource languages in particular, with 200 training instances in the target or related languages producing better results on dependency parsing than a model trained on all available data. \citet{lauscher2020} further show that as few as 10 samples in the target language can double parsing performance. \citet{ustun-etal-2020-udapter} propose UDapters, which integrate language and task-specific adaptation modules into the parser to improve cross-lingual, zero-shot performance.

Considering factors complementary to language is equally important: MLMs can for instance be improved for specific domains such as Twitter or medical texts by fine-tuning on the same or related sources \citep{dai2020, gururangan2020}. For dependency parsing, the use of data from matching genres has been explored by \citet{plank2011}, who find improvements for English and Dutch. This is further confirmed for German by \citet{rehbein-bildhauer-2017-data}.

Automatically inferred topics \citep{ruder2017} as well as more abstract selection criteria such as overlapping part-of-speech sequences \citep{sogaard2011, rosa2015} have also proven effective at selecting syntactically similar training instances. \citet{vania2019} further demonstrate that when word embeddings of mutually unintelligible languages align with respect to POS, cross-lingual transfer remains especially effective. With respect to data-driven domain representations, ~\citet{stymne-2020-cross} shows that treebank embeddings can be used to successfully transfer knowledge from in-domain cross-lingual source treebanks when used in conjunction with in-language, out-of-domain data. In this work, we will rely solely on treebank genre labels as weak supervision and forgo the use of in-language training data as well as instance-level annotations thereof (e.g.\ POS tags).

Recently, contextualized embeddings have been shown to contain useful information for training data selection. \citet{aharoni2020} find that clusters formed by embeddings from untuned, monolingual language models correspond well to the genres of their five-domain corpus. Training an English-to-German machine translation model on only the closest embedded sentences to their target 2k-sentence development set outperformed a model trained on the entire dataset.

Although all aforementioned methods assume some degree of in-language training data, our methods will not have access to any annotated target data and will be trained exclusively on out-of-language instances. Building on information stored in pre-trained contextual embeddings, we extend genre-based data selection into the massively multilingual, 104-language, 18-genre setting of Universal Dependencies \citep{ud27}. While previous work further assumed sentence-level genre labels \citep{ruder2017,aharoni2020}, our methods will only have access to treebank-level metadata. An instance's genre will therefore have to be inferred using weakly supervised approaches. To the best of our knowledge, this constitutes the first application of UD's instance-level genre distribution to the selection of training data for zero-shot, cross-lingual dependency parsing.

\section{Genre in Universal Dependencies}\label{sec:ud-genre}

\begin{figure}
    \centering
    \pdftooltip{\hspace{-.5em}\includegraphics[width=.49\textwidth]{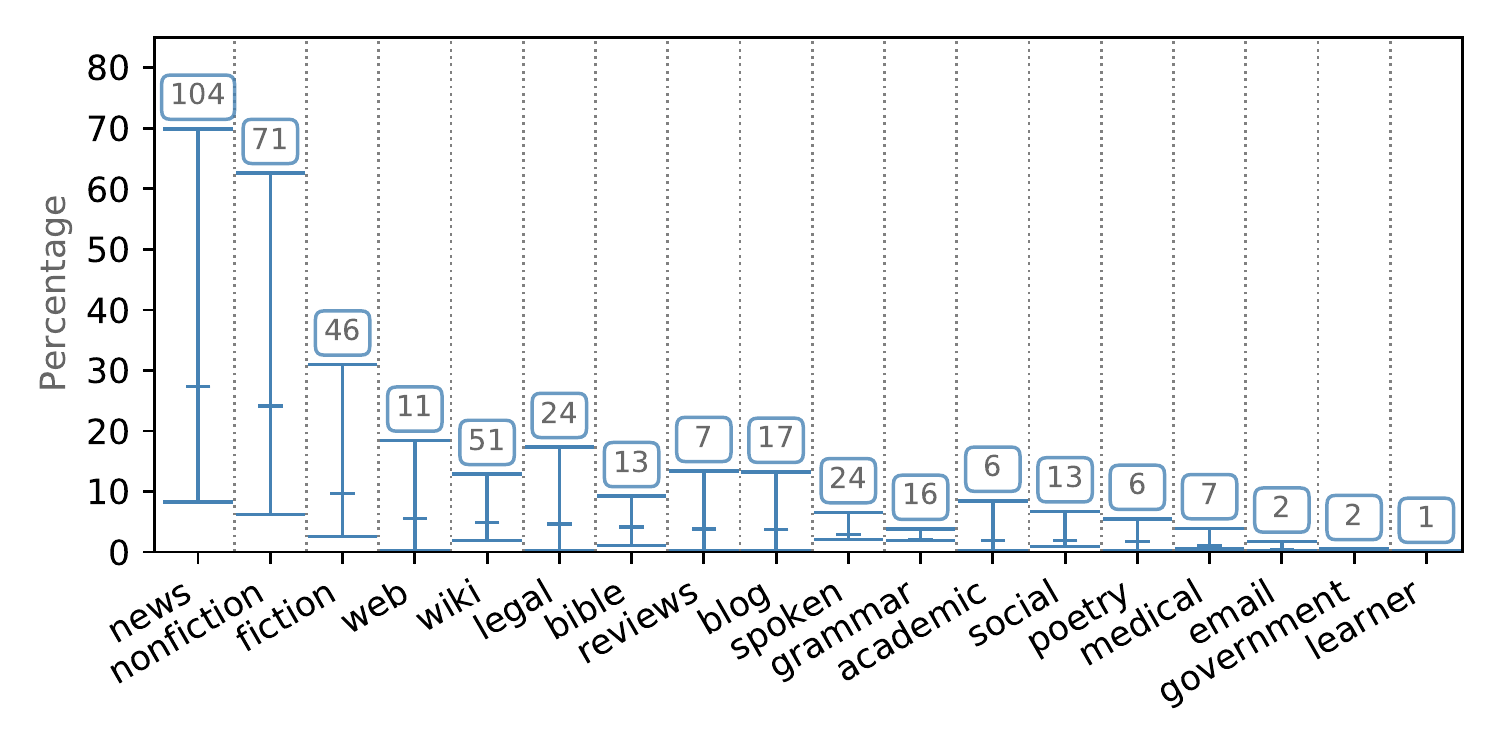}}{Screenreader Caption: news: part of 104 treebanks, 8\% to 70\% of sentences (27\% under uniformity). nonfiction: part of 71 treebanks, 6\% to 63\% of sentences (24\% under uniformity). fiction: part of 46 treebanks, 2\% to 31\% of sentences (10\% under uniformity). web: part of 11 treebanks, less than 1\% to 18\% of sentences (6\% under uniformity). wiki: part of 51 treebanks, 2\% to 13\% of sentences (5\% under uniformity). legal: part of 24 treebanks, less than 1\% to 17\% of sentences (5\% under uniformity). bible: part of 13 treebanks, 1\% to 9\% of sentences (4\% under uniformity). reviews: part of 7 treebanks, less than 1\% to 13\% of sentences (4\% under uniformity). blog: part of 17 treebanks, less than 1\% to 13\% of sentences (4\% under uniformity). spoken: part of 24 treebanks, 2\% to 7\% of sentences (3\% under uniformity). grammar: part of 16 treebanks, 2\% to 4\% of sentences (2\% under uniformity). academic: part of 6 treebanks, less than 1\% to 8\% of sentences (2\% under uniformity). social: part of 13 treebanks, 1\% to 7\% of sentences (2\% under uniformity). poetry: part of 6 treebanks, less than 1\% to 5\% of sentences (2\% under uniformity). medical: part of 7 treebanks, less than 1\% to 4\% of sentences (1\% under uniformity). email: part of 2 treebanks, less than 1\% to 2\% of sentences (less than 1\% under uniformity). government: part of 2 treebanks, less than 1\% to less than 1\% of sentences (less than 1\% under uniformity). learner: part of 1 treebank, less than 1\% to less than 1\% of sentences (less than 1\% under uniformity).}
    \caption{\textbf{Genre Distribution in UD.} Ranges indicate upper/lower bounds for sentences per genre inferred from UD metadata. Center marker reflects the distribution under the assumption that genres within treebanks are uniformly distributed. Labels above the bars indicate the number of treebanks which contain each genre.}
    \label{fig:ud-genre-dist}
\end{figure}

Universal Dependencies \citep{nivre-etal-2016-universal} offer annotations for a broad spectrum of languages, with 104 in version 2.7~\cite{ud27}. Of the 1.38 million sentences from the 177 treebanks which we consider, 64 are test-set only and many in this latter third constitute the sole treebank of the language they are in. Such data sparsity becomes even more critical when both the language and the domain are highly specialized and under-resourced.

As more low-resource languages are added in this manner and as the vast majority of the world's languages remain without annotated data, it becomes important to consider new signals for selecting training data in zero-shot scenarios. If no data in the target language are available, we hypothesize that characteristics of most genres are stable enough across languages to offer a useful guiding criterion for data selection in cross-lingual dependency parsing.

For 26 of the 177 treebanks, their authors have provided sentence-level genre labels. However, these annotations cover only 6\% of UD sentences and are typically incompatible across treebanks (with few exceptions such as PUD). At the treebank level, UD fortunately provides 18 approximated genre labels: \textit{academic, bible, blog, email, fiction, government, grammar-examples, learner-essays, legal, medical, news, nonfiction, poetry, reviews, social, spoken, web, wiki}.

Genres such as \textit{wiki} likely have stronger internal consistency due to cross-lingual creation guidelines. Others such as \textit{fiction} or \textit{web} may have higher variance. While these UD-provided labels are far from perfectly defined \citep{nivre-etal-2020-universal}, they nonetheless allow us to operationalize our hypothesis: If genre is globally consistent, it must have a positive effect on cross-lingual transfer performance.

From Figure \ref{fig:ud-genre-dist} it is evident that these genres are heavily imbalanced. The minimum number of sentences in a genre is inferred from the sum over the number of instances in treebanks containing only that genre. The upper bound is the sum of all treebanks containing the genre among others. As indicated by these distributional bounds, news articles may constitute up to 70\% of the whole UD dataset. Even assuming uniform genre distributions within each treebank (center marker), over half of all sentences in UD would fall into either the \textit{news} or the \textit{non-fiction} category.

Genres with highly specific lexical and/or structural features such as \textit{spoken}, \textit{social} or \textit{medical} are much more underrepresented. Furthermore, they are often only a small part of larger genre mixtures (117 treebanks include multiple genre-labels). These mixtures, with up to 10 genres in one treebank, may contain related genres (e.g.\ \textit{news}, \textit{non-fiction}, \textit{web}), but also unrelated ones (e.g.\ \textit{medical}, \textit{poetry}, \textit{social}, \textit{web}) depending on what data was available to authors during annotation.

Out-of-the-box, treebank-level genre labels appear to be highly noisy (see also \citealp{nivre-etal-2020-universal}). Additionally, individual treebanks are labeled with multiple genres while lacking such labels at the sentence level. We hypothesize that it is therefore necessary to predict \textit{instance-level} genre distributions before targeted data selection can be effective.

\section{Targeted Data Selection}\label{sec:method}

In order to measure the effect of genre on the targeted selection of training data, we depart from previous treebank-level selection (Section \ref{sec:relwork}) and introduce three new types of instance-level selection strategies in the following section. They are evaluated on the task of zero-shot dependency parsing in Sections \ref{sec:experiments} and \ref{sec:results}. All of them build on contextualized embeddings learned by the mBERT \citep{devlin2019} masked language model (MLM). While MLMs still lack the full breadth of the languages covered in UD (mBERT covers 56 of the 104 languages), they have proven robust in zero-shot scenarios \citep{devlin2019, brown2020} and have also been found to contain a certain amount of genre information --- at least monolingually (\citealp{aharoni2020}; Section~\ref{sec:relwork}). We evaluate whether UD's definition of genre is also recoverable from these data-driven representations and whether these categories hold cross-lingually.

\subsection{Closest Sentence Selection}\label{sec:method-sentence}
\paragraph{\textsc{Sent}}
Akin to the strategy used by \citet{aharoni2020}, this \textsc{Sentence}-based method attempts to find the most relevant training data by computing the mean embedding of $n$ unannotated target data samples and retrieving the top-$k$ closest non-target instances according to their cosine distance in embedding space. Notable differences from their original method are the use of a much smaller target data sample ($n=100$ versus $n=2000$) as well as the use of mBERT instead of English-only BERT embeddings \citep{devlin2019} due to our cross-lingual setting.

While the monolingual BERT embeddings were found to represent genre to some degree, such MLM embeddings likely contain many more dimensions of semantic and syntactic information. The \textsc{Sent} method alone is therefore not guaranteed to represent data selection by genre as stronger factors may override these signals. Additionally, \citet{aharoni2020}'s setup assumed five clearly-defined genres with instance-level annotations while UD has 18 genres with varying degrees of specificity which are only defined in the treebank-level metadata.

\subsection{Genre Selection}\label{sec:method-genre}
\paragraph{\textsc{Meta}}
Separately to MLM embedding-based selection, we evaluate the effectiveness of using the manually assigned genre labels listed in each treebank's metadata. As seen in Section \ref{sec:ud-genre}, these labels can be noisy and have variable interpretations across treebanks. Furthermore, each treebank is assigned up to 10 genres, making instance-level selection as in the previous method impossible.

\paragraph{\textsc{\boot}}
To bridge this gap to sentence-level selection, we introduce a bootstrapping procedure which iteratively learns an instance-level classifier for UD genre. Each sentence is encoded through mBERT's \texttt{CLS} token before passing to a classification layer. The model is initialized using standard mBERT weights and begins by training on single-genre treebanks (i.e.\ standard supervised learning). It then predicts sentence labels for treebanks containing these initial genres. Above a prediction threshold of $0.99 \in [0, 1]$, these are added as new training data for the next round of training. When only one unclassified genre remains in a treebank, all remaining instances are inferred to be of that last genre. Using this procedure, a single genre label is assigned to each sentence in UD within three steps.

\paragraph{} Compared to closest sentence selection (\textsc{Sent}), both of the former methods have the added benefit that no target-data is required in order to make the final training data selection. The training corpus simply consists of all instances labelled as belonging to a genre (\boot) or to a treebank containing the genre in question (\textsc{Meta}).

%
%

\begin{table*}
\centering
\resizebox{\textwidth}{!}{
\begin{tabular}{llllcrl}
\toprule
\textsc{Target} & \textsc{Authors} & \textsc{Language} & \textsc{Family} & \textsc{mB} & \textsc{Size} & \textsc{Genre}\\
\midrule
SWL-SSLC & \citet{swl-sslc} & Swedish Sign Language & Signed Language & $\times$ & 203 & {\def\svgwidth{.65em}
\begingroup%
  \makeatletter%
  \providecommand\color[2][]{%
    \errmessage{(Inkscape) Color is used for the text in Inkscape, but the package 'color.sty' is not loaded}%
    \renewcommand\color[2][]{}%
  }%
  \providecommand\transparent[1]{%
    \errmessage{(Inkscape) Transparency is used (non-zero) for the text in Inkscape, but the package 'transparent.sty' is not loaded}%
    \renewcommand\transparent[1]{}%
  }%
  \providecommand\rotatebox[2]{#2}%
  \newcommand*\fsize{\dimexpr\f@size pt\relax}%
  \newcommand*\lineheight[1]{\fontsize{\fsize}{#1\fsize}\selectfont}%
  \ifx\svgwidth\undefined%
    \setlength{\unitlength}{384bp}%
    \ifx\svgscale\undefined%
      \relax%
    \else%
      \setlength{\unitlength}{\unitlength * \real{\svgscale}}%
    \fi%
  \else%
    \setlength{\unitlength}{\svgwidth}%
  \fi%
  \global\let\svgwidth\undefined%
  \global\let\svgscale\undefined%
  \makeatother%
  \begin{picture}(1,1)%
    \lineheight{1}%
    \setlength\tabcolsep{0pt}%
    \put(0,0){\includegraphics[width=\unitlength,page=1]{img/icon-spoken.pdf}}%
  \end{picture}%
\endgroup%
}{} spoken\\
SA-UFAL & \citet{sa-ufal} & Sanskrit & Indo-European & $\times$ & 230 & {\def\svgwidth{.6em}
\begingroup%
  \makeatletter%
  \providecommand\color[2][]{%
    \errmessage{(Inkscape) Color is used for the text in Inkscape, but the package 'color.sty' is not loaded}%
    \renewcommand\color[2][]{}%
  }%
  \providecommand\transparent[1]{%
    \errmessage{(Inkscape) Transparency is used (non-zero) for the text in Inkscape, but the package 'transparent.sty' is not loaded}%
    \renewcommand\transparent[1]{}%
  }%
  \providecommand\rotatebox[2]{#2}%
  \newcommand*\fsize{\dimexpr\f@size pt\relax}%
  \newcommand*\lineheight[1]{\fontsize{\fsize}{#1\fsize}\selectfont}%
  \ifx\svgwidth\undefined%
    \setlength{\unitlength}{336bp}%
    \ifx\svgscale\undefined%
      \relax%
    \else%
      \setlength{\unitlength}{\unitlength * \real{\svgscale}}%
    \fi%
  \else%
    \setlength{\unitlength}{\svgwidth}%
  \fi%
  \global\let\svgwidth\undefined%
  \global\let\svgscale\undefined%
  \makeatother%
  \begin{picture}(1,1.14285714)%
    \lineheight{1}%
    \setlength\tabcolsep{0pt}%
    \put(0,0){\includegraphics[width=\unitlength,page=1]{img/icon-fiction.pdf}}%
  \end{picture}%
\endgroup%
}{} fiction\\
KPV-Lattice & \citet{kpv-tbs} & Komi Zyrian & Uralic & $\times$ & 435 & {\def\svgwidth{.6em}}{} fiction\\
TA-TTB & \citet{ta-ttb} & Tamil & Dravidian & \checkmark & 600 & {\def\svgwidth{.7em}
\begingroup%
  \makeatletter%
  \providecommand\color[2][]{%
    \errmessage{(Inkscape) Color is used for the text in Inkscape, but the package 'color.sty' is not loaded}%
    \renewcommand\color[2][]{}%
  }%
  \providecommand\transparent[1]{%
    \errmessage{(Inkscape) Transparency is used (non-zero) for the text in Inkscape, but the package 'transparent.sty' is not loaded}%
    \renewcommand\transparent[1]{}%
  }%
  \providecommand\rotatebox[2]{#2}%
  \newcommand*\fsize{\dimexpr\f@size pt\relax}%
  \newcommand*\lineheight[1]{\fontsize{\fsize}{#1\fsize}\selectfont}%
  \ifx\svgwidth\undefined%
    \setlength{\unitlength}{432bp}%
    \ifx\svgscale\undefined%
      \relax%
    \else%
      \setlength{\unitlength}{\unitlength * \real{\svgscale}}%
    \fi%
  \else%
    \setlength{\unitlength}{\svgwidth}%
  \fi%
  \global\let\svgwidth\undefined%
  \global\let\svgscale\undefined%
  \makeatother%
  \begin{picture}(1,0.88888889)%
    \lineheight{1}%
    \setlength\tabcolsep{0pt}%
    \put(0,0){\includegraphics[width=\unitlength,page=1]{img/icon-news.pdf}}%
  \end{picture}%
\endgroup%
}{} news\\
GL-TreeGal & \citet{gl-treegal} & Galician & Indo-European & \checkmark & 1,000 & {\def\svgwidth{.7em}}{} news\\
YUE-HK & \citet{yue-hk} & Cantonese & Sino-Tibetan & $\times$ & 1,004 & {\def\svgwidth{.65em}}{} spoken\\
CKT-HSE & \citet{ckt-hse} & Chukchi & Chukotko-Kamchatkan & $\times$ & 1,004 & {\def\svgwidth{.65em}}{} spoken\\
FO-OFT & \citet{fo-oft} & Faroese & Indo-European & $\times$ & 1,208 & {\def\svgwidth{.7em}
\begingroup%
  \makeatletter%
  \providecommand\color[2][]{%
    \errmessage{(Inkscape) Color is used for the text in Inkscape, but the package 'color.sty' is not loaded}%
    \renewcommand\color[2][]{}%
  }%
  \providecommand\transparent[1]{%
    \errmessage{(Inkscape) Transparency is used (non-zero) for the text in Inkscape, but the package 'transparent.sty' is not loaded}%
    \renewcommand\transparent[1]{}%
  }%
  \providecommand\rotatebox[2]{#2}%
  \newcommand*\fsize{\dimexpr\f@size pt\relax}%
  \newcommand*\lineheight[1]{\fontsize{\fsize}{#1\fsize}\selectfont}%
  \ifx\svgwidth\undefined%
    \setlength{\unitlength}{480bp}%
    \ifx\svgscale\undefined%
      \relax%
    \else%
      \setlength{\unitlength}{\unitlength * \real{\svgscale}}%
    \fi%
  \else%
    \setlength{\unitlength}{\svgwidth}%
  \fi%
  \global\let\svgwidth\undefined%
  \global\let\svgscale\undefined%
  \makeatother%
  \begin{picture}(1,0.8)%
    \lineheight{1}%
    \setlength\tabcolsep{0pt}%
    \put(0,0){\includegraphics[width=\unitlength,page=1]{img/icon-wiki.pdf}}%
  \end{picture}%
\endgroup%
}{} wiki\\
TE-MTG & \citet{te-mtg} & Telugu & Dravidian & \checkmark & 1,328 & {\def\svgwidth{.6em}
\begingroup%
  \makeatletter%
  \providecommand\color[2][]{%
    \errmessage{(Inkscape) Color is used for the text in Inkscape, but the package 'color.sty' is not loaded}%
    \renewcommand\color[2][]{}%
  }%
  \providecommand\transparent[1]{%
    \errmessage{(Inkscape) Transparency is used (non-zero) for the text in Inkscape, but the package 'transparent.sty' is not loaded}%
    \renewcommand\transparent[1]{}%
  }%
  \providecommand\rotatebox[2]{#2}%
  \newcommand*\fsize{\dimexpr\f@size pt\relax}%
  \newcommand*\lineheight[1]{\fontsize{\fsize}{#1\fsize}\selectfont}%
  \ifx\svgwidth\undefined%
    \setlength{\unitlength}{384bp}%
    \ifx\svgscale\undefined%
      \relax%
    \else%
      \setlength{\unitlength}{\unitlength * \real{\svgscale}}%
    \fi%
  \else%
    \setlength{\unitlength}{\svgwidth}%
  \fi%
  \global\let\svgwidth\undefined%
  \global\let\svgscale\undefined%
  \makeatother%
  \begin{picture}(1,1)%
    \lineheight{1}%
    \setlength\tabcolsep{0pt}%
    \put(0,0){\includegraphics[width=\unitlength,page=1]{img/icon-grammar.pdf}}%
  \end{picture}%
\endgroup%
}{} grammar\\
MYV-JR & \citet{myv-jr} & Erzya & Uralic & $\times$ & 1,690 & {\def\svgwidth{.6em}}{} fiction\\
QHE-HIENCS & \citet{qhe-hiencs} & Hindi-English &  Code-Switched & $\sim$ & 1,800 & {\def\svgwidth{.6em}
\begingroup%
  \makeatletter%
  \providecommand\color[2][]{%
    \errmessage{(Inkscape) Color is used for the text in Inkscape, but the package 'color.sty' is not loaded}%
    \renewcommand\color[2][]{}%
  }%
  \providecommand\transparent[1]{%
    \errmessage{(Inkscape) Transparency is used (non-zero) for the text in Inkscape, but the package 'transparent.sty' is not loaded}%
    \renewcommand\transparent[1]{}%
  }%
  \providecommand\rotatebox[2]{#2}%
  \newcommand*\fsize{\dimexpr\f@size pt\relax}%
  \newcommand*\lineheight[1]{\fontsize{\fsize}{#1\fsize}\selectfont}%
  \ifx\svgwidth\undefined%
    \setlength{\unitlength}{336bp}%
    \ifx\svgscale\undefined%
      \relax%
    \else%
      \setlength{\unitlength}{\unitlength * \real{\svgscale}}%
    \fi%
  \else%
    \setlength{\unitlength}{\svgwidth}%
  \fi%
  \global\let\svgwidth\undefined%
  \global\let\svgscale\undefined%
  \makeatother%
  \begin{picture}(1,1.14285714)%
    \lineheight{1}%
    \setlength\tabcolsep{0pt}%
    \put(0,0){\includegraphics[width=\unitlength,page=1]{img/icon-social.pdf}}%
  \end{picture}%
\endgroup%
}{} social\\
QTD-SAGT & \citet{qtd-sagt} & Turkish-German & Code-Switched & $\sim$ & 1,891 & {\def\svgwidth{.65em}}{} spoken\\
\bottomrule
\end{tabular}
}
\caption{\label{tab:target-tbs} \textbf{Target Treebanks} with language family (\textsc{Family}), inclusion in mBERT pre-training (\textsc{mB}; included (\checkmark), excluded ($\times$), highly-related languages included ($\sim$)), total number of sentences (\textsc{Size}) and UD-provided \textsc{Genre}.}
\end{table*}

%
%

\subsection{Closest Cluster Selection}\label{sec:method-cluster}
\paragraph{\textsc{GMM}}
As shown by \citet{aharoni2020}, monolingual BERT embeddings can be clustered into distinct domains using common clustering algorithms such as Gaussian Mixture Models (GMMs). Using mBERT embeddings, we evaluate whether this holds cross-lingually by clustering each treebank into the number of genres which it is said to contain according to the UD-provided metadata.
Deviating from previous work, which only uses these clusters for preliminary analyses, we then use them directly for data selection. By computing a mean embedding for each cluster and choosing the closest one to the mean target sample embedding (same as \textsc{Sent}), the most similar data is selected in bulk from each treebank. By only selecting clusters from treebanks for which the metadata states that the target genre is contained, this allows us to identify clusters which most likely correspond to the target genre while avoiding the manual labelling of clusters across 104 languages.

\paragraph{\textsc{LDA}}
We also evaluate a clustering method based purely on lexical features (i.e.\ n-grams) instead of pre-trained contextual embeddings. While the selection of the most relevant cluster from each treebank is performed using the same mean embedding distance methodology as for \textsc{GMM}, we use Latent Dirichlet Allocation (\citealp{blei2003}; LDA) for the initial clustering step. This decouples the genre-segmentation step from the multitude of non-genre dimensions in the embeddings themselves, while simultaneously not relying on LDA alone for the final data selection (as in \citealp{plank2011,mukherjee-etal-2017-creating}). Furthermore, this setup allows us to extract genres from languages and scripts unknown to mBERT as well as to compare whether the GMM clusters correspond to those found by using surface-level lexical information alone.

\section{Experimental Setup}\label{sec:experiments}

\subsection{Target Treebanks}\label{sec:target-tbs}

We evaluate the effect of genre on training data selection using a set of 12 target treebanks from the low-resource end of UD. For our purposes, low-resource is defined as treebanks with more than 200 and less than 2,000 sentences in total and with fewer than 5,000 in-language sentences in UD.

In order to distinguish the effects of genre specifically, we only use single-genre target treebanks and leave the investigation of genre-mixtures to future work. As seen in Table \ref{tab:target-tbs}, the final set of targets is diverse with respect to genre, language family and their availability during mBERT pre-training.

Only three of the target languages are included in mBERT pre-training, with seven not being covered at all and two having strongly related languages in mBERT's repertoire: Hindi-English (QHE) $\rightarrow$ Hindi, English as well as Turkish-German (QTD) $\rightarrow$ Turkish, German.

The six included genres cover the high-resource \textit{news} ({\def\svgwidth{.7em}}) and \textit{fiction} ({\def\svgwidth{.6em}}) as well as the medium resource \textit{wiki} ({\def\svgwidth{.7em}}) and the lower resource \textit{spoken} ({\def\svgwidth{.65em}}), \textit{grammar-examples} ({\def\svgwidth{.6em}}) and \textit{social} ({\def\svgwidth{.6em}}).

\subsection{Data Selection Setup}\label{sec:data-setup}

%
%

\newcommand{\sigdagger}{\raisebox{.15em}{\footnotesize\textdagger}}
\newcommand{\sigdiamond}{\raisebox{.25em}{\footnotesize$\diamond$}}
\begin{table*}
\centering
\resizebox{\textwidth}{!}{
\begin{tabular}{lrrrrrrrrrrrrr}
\toprule
\textsc{Setup} & \multicolumn{1}{c}{\textsc{SWL} {\def\svgwidth{.65em}}} & \multicolumn{1}{c}{\textsc{SA} {\def\svgwidth{.6em}}} & \multicolumn{1}{c}{\textsc{KPV} {\def\svgwidth{.6em}}} & \multicolumn{1}{c}{\textsc{TA} {\def\svgwidth{.7em}}} & \multicolumn{1}{c}{\textsc{GL} {\def\svgwidth{.7em}}} & \multicolumn{1}{c}{\hspace{-.1em}\textsc{YUE} {\def\svgwidth{.65em}}} & \multicolumn{1}{c}{\hspace{-.1em}\textsc{CKT} {\def\svgwidth{.65em}}} & \multicolumn{1}{c}{\textsc{FO} {\def\svgwidth{.7em}}} & \multicolumn{1}{c}{\textsc{TE} {\def\svgwidth{.6em}}} & \multicolumn{1}{c}{\hspace{-.5em}\textsc{MYV} {\def\svgwidth{.6em}}} & \multicolumn{1}{c}{\hspace{-.1em}\textsc{QHE} {\def\svgwidth{.6em}}} & \multicolumn{1}{c}{\textsc{QTD} {\def\svgwidth{.65em}}} & \multicolumn{1}{|c}{\textsc{AVG}} \\
\midrule
\textsc{Target} & 28.01 & 15.74 & 13.36 & 64.05 & 80.94 & --- & --- & 49.55 & 83.63 & --- & 62.66 & 55.04 & \multicolumn{1}{|r}{50.28}\\
\midrule[.3pt]
\textsc{Rand} & 3.67 & \textbf{24.81} & 10.88 & 50.73 & 77.65 & 33.31 & 15.54 & 61.88 & 67.68 & 20.01 & \textbf{27.01} & 44.57 & \multicolumn{1}{|r}{36.48} \\
\textsc{Sent} & 3.55 & 23.72 & 13.71 & 47.93 & 77.55 & 35.78 & 16.44 & 62.49 & 68.05 & \textbf{22.90} & 26.46 & 42.74 & \multicolumn{1}{|r}{36.78} \\
\midrule[.3pt]
\midrule[.3pt]
\textsc{Meta} & 6.50 & 24.29 & 10.22 & 50.43 & 76.63 & 31.19 & 11.62 & 61.23 & 64.91 & 20.41 & 9.42 & 42.58 & \multicolumn{1}{|r}{34.12} \\
\midrule[.3pt]
\textsc{Boot} & 5.20 & 21.80 & \sigdagger21.09 & 49.43 & 76.66 & \sigdagger49.85 & 18.40 & \sigdagger66.25 & 65.56 & 19.46 & 14.75 & 43.80 & \multicolumn{1}{|r}{37.69} \\
\midrule[.3pt]
\textsc{GMM} & 4.85 & 22.93 & \sigdagger20.91 & \sigdagger\textbf{51.53} & \textbf{77.75} & \sigdagger\textbf{49.92} & \sigdagger\textbf{19.81} & \sigdagger68.25 & 67.87 & 20.15 & 15.09 & \textbf{45.38} & \multicolumn{1}{|r}{38.70} \\
\textsc{LDA} & \textbf{6.62} & 23.70 & \sigdagger\textbf{22.27} & 49.17 & 77.01 & \sigdagger49.40 & \sigdagger19.05 & \sigdagger\textbf{68.29} & \sigdagger\textbf{68.56} & 20.54 & 15.16 & 44.72 & \multicolumn{1}{|r}{\textbf{38.71}} \\
\bottomrule
\end{tabular}
}
\caption{\label{tab:parsing-results} \textbf{Zero-shot Parsing Results.} LAS for test splits of target treebanks using training data from target/proxy in-language treebanks (\textsc{Target}; where available), random sentence selection (\textsc{Rand}), closest sentence selection (\textsc{Sent}), treebanks containing target genre (\textsc{Meta}), instances classified as target genre (\textsc{Boot}) and closest cluster selection (\textsc{GMM} and \textsc{LDA}). Scores marked with \sigdagger{} significantly outperform  \textsc{Target}, \textsc{Rand}, \textsc{Sent} and \textsc{Meta}.}
\end{table*}

%
%

In order to train parsers for these largely test-only treebanks, we compare seven proxy training data selection strategies for each target (note that only the first strategy uses in-language training data).

\paragraph{\textsc{Target}} Where available, we use the true target training split as a performance upper bound against which to compare our methods. These are available for the six targets: SWL-SSLC, TA-TTB, GL-TreeGal, TE-MTG, QHE-HIENCS and QTD-SAGT. For three targets without training splits, we make use of proxy in-language data: SA-Vedic \citep{sa-vedic} for SA-UFAL, KPV-IKDP  \citep{kpv-tbs} for KPV-Lattice and FO-FarPaHC \citep{fo-farpahc} for FO-OFT. For the targets YUE-HK, CKT-HSE and MYV-JR no in-language training data are currently available.

\paragraph{\textsc{Rand}} selects a random sample of $n_{\text{rand}}$ sentences from the non-target-language UD. We do not restrict this selection to treebanks containing the target genre such that data from a more diverse pool of languages may be selected. To ensure an equivalent comparison, we set $n_{\text{rand}}$ to the mean of the number of instances selected by \boot, \textsc{LDA} and \textsc{GMM} (see Appendix \ref{sec:additional-results} for values of $n_{\text{rand}}$).

\paragraph{\textsc{Sent}} selection (see Section \ref{sec:method-sentence}) is based on the mean embedding of 100 target sentences and retrieves the top-$k$ closest out-of-language sentences from all of UD independently of genre. Since $k$ needs to be chosen manually, we set it to the number of instances selected by \textsc{GMM}, which is equally dependent on mBERT embeddings.

\paragraph{\textsc{Meta}} selects all non-target language treebanks which are denoted to contain the target genre (i.e.\ both single-genre treebanks as well as mixtures). These data pools make up the largest training corpora in our setup (up to 524k instances for \textit{news}) and also subsume the other genre-based selection methods \boot, \textsc{LDA} and \textsc{GMM}. In this way, it acts as an upper bound in terms of data quantity as well as a baseline for whether treebank-level metadata alone can aid data selection.

\paragraph{\textsc{Boot}} selects only the specific instances classified as being in the target genre for use as training data. The classifier is trained according to the bootstrapping method outlined in Section \ref{sec:method-genre}. In order to avoid the memorization of target data, we exclude all data in the target languages from the classifier training process.

\paragraph{\textsc{GMM}} clusters each treebank into the number of genres denoted by its metadata using mean-pooled mBERT embeddings for each sentence. Training data is then selected according to the closest-cluster procedure outlined in Section \ref{sec:method-cluster}.

\paragraph{\textsc{LDA}} works analogously to \textsc{GMM}, but uses LDA to cluster sentences. It uses bags of character 3--6-grams and no language-specific resources (e.g.\ stop word lists) in order to remain as cross-lingually comparable as possible. Hyperparameters were tuned as outlined in Section \ref{sec:training-setup}.

\paragraph{} All methods relying on unannotated target data for the data selection process use 100 random sentences from the target treebank (changes across random initializations). In practical terms, this corresponds to having access to a small amount of target-like data --- without gold dependency structures --- and selecting the best possible training data for which we do have annotations.

Alternatively, \boot{} (as well as \textsc{Meta} and \textsc{Rand} implicitly) work in a fully zero-shot manner as we only assume knowledge of the intended target genre, but do not assume access to the target sentences \textit{nor} their annotations.

\subsection{Training Setup}\label{sec:training-setup}

We use the biaffine attention parser~\cite{dozat2017} implementation of MaChAmp v0.2~\citep{machamp} with default hyperparameters. Each step involving non-deterministic components is rerun using three random seeds.

For efficiency reasons, the seven largest treebanks were subsampled to 20k instances per split. Performance is measured using the labeled attachment scores (LAS) averaged across random initializations. Additionally, we report unlabeled attachment scores (UAS), the number of selected instances as well as the variance across runs in Appendix \ref{sec:additional-results}. Significance is evaluated at $\alpha < 0.05$ using a paired bootstrapped sign test with 10k resampling and Bonferroni correction \citep{bonferroni1936} for the multiple comparisons across random initializations. Appendix \ref{sec:training-details} lists all additional hyperparameter settings.

It is important to note that besides the upper bound in-language setup (\textsc{Target}), no parser is trained on in-language data. For the tuning of method-specific hyperparameters (\textsc{LDA} features, \textsc{Boot} thresholds), development sets of the five treebanks containing such splits were used: SWL-SSLC, TA-TTB, TE-MTG, QHE-HIENCS and QTD-SAGT (details in Appendix \ref{sec:training-details}). During parser training, development data for early stopping is based solely on the out-of-language data selected by each method and not on the in-language target data itself (also excluding constituent languages for code switched targets). Results are reported on each target's test set without any further tuning.

\section{Results}\label{sec:results}

\subsection{Zero-shot Parsing Results}\label{sec:parsing-results}

As expected, Table \ref{tab:parsing-results} shows that training the parser on target data (\textsc{Target}) results in the best overall performance even though the training corpora for these setups almost never exceed 1k instances. The target treebanks for which in-language data are available, consolidate into a final average of 50.28 LAS. This highlights the overall difficulty of parsing these low-resource treebanks. As the parser is initialized using mBERT, the scores on Tamil (TA), Galician (GL) and Telugu (TE), which are included in its pre-training, are highest overall compared to non-included languages or code-switched variants.

It is noteworthy that when a same-language proxy treebank was used for parser training, scores are lower compared to the other methods. In these three cases, namely Sanskrit (SA), Komi Zyrian (KPV) and Faroese (FO), none of the proxy treebanks include the target's genre which may be a strong contributing factor to this discrepancy.

Turning to our zero-shot setups, \textsc{Meta} data selection based on treebank-level annotations alone performs worst overall at 34.12 LAS despite constituting the largest training corpora in each setup (see Appendix \ref{sec:additional-results} for details). Compared to the \textsc{Target} upper bounds, it shows how training on two orders of magnitudes more data can still be insufficient if they do not follow the target distribution.

Both \textsc{Rand} and \textsc{Sent} outperform the \textsc{Meta} baseline at 36.48 and 36.78 LAS respectively. These aggregated scores also highlight that sentence-based selection alone insufficiently captures cross-lingual characteristics as to outperform random chance in most cases.

In contrast, combining latent information in the MLM embeddings with higher-level genre information leads to performance increases not achievable by either method alone. Both \textsc{GMM} and \textsc{LDA} achieve the highest scores across the majority of target treebanks and the highest cross-lingual averages of 38.70 LAS and 38.71 LAS respectively. These scores reflect their similar performance across targets, however we do observe that \textsc{LDA} achieves slightly higher scores on languages which are not included in mBERT pre-training: e.g.\ Swedish Sign Language (SWL), Sanskrit (SA) and Komi Zyrian (KPV). We hypothesize that this is a result of \textsc{GMM}'s dependence on latent information in the mBERT embeddings while \textsc{LDA} constructs clusters independently, based solely on surface-level lexical features (i.e.\ n-grams).

Finally, amplifying genre information in the mBERT embeddings using our \boot{} method also leads to performance increases compared to using untuned embeddings or the coarser grained treebank-level metadata. While it does not entirely reach the performance of the cluster selection methods, its overall average of 37.69 LAS as well as generally similar performance patterns to \textsc{LDA} and \textsc{GMM} lead us to believe that all three methods are picking up on and are amplifying similar latent genre information. As an added benefit, \boot{} is able to reach this competitive performance without the need for any target data samples (as opposed to \textsc{GMM} and \textsc{LDA} which use 100 raw samples for cluster selection).

Using our proposed genre-based selection methods we are therefore able to consistently outperform in-language/out-of-genre upper bounds for these low-resource target treebanks. Comparing our results to \citet{machamp} who train an identical parser architecture on each UD treebank's respective training split, proxy treebank (for test-only) or all of UD, our methods significantly outperform their best models on five of twelve target treebanks.\footnote{We compare against the highest score across all of their proposed models for each treebank.} There are significant increases for both SA-UFAL (16.5 $\rightarrow$ 23.7 LAS) and KPV-Lattice (11.7 $\rightarrow$ 22.3 LAS).\footnote{\citet{dehouck2019} achieve higher scores using a parsing architecture with POS and morphological features.} For the targets YUE-HK (32.7 $\rightarrow$ 49.9 LAS), CKT-HSE (15.3 $\rightarrow$ 19.8 LAS) and FO-OFT (62.7 $\rightarrow$ 68.3 LAS), these scores furthermore constitute --- to the best of our knowledge --- state-of-the-art results without requiring annotated in-language data.

\subsection{Analysis of Selected Data}\label{sec:analysis}

Further analyzing the patterns of data selection allows us to identify some of the reasons behind the differences in performance (visualizations can be found in Appendix \ref{sec:selection-analysis}).

\textsc{Rand} closely follows the overall data distribution in UD, selecting the most instances from the largest treebanks such as German-HDT \citep{de-hdt} and selecting none to almost none from low-resource treebanks. \textsc{Sent} follows a similar distribution albeit rarely selecting zero instances from any given language. This behaviour does not change substantially between targets, indicating less targeted data selection.

While the larger language diversity of the aforementioned \textsc{Rand} and \textsc{Sent} does not seem to be enough to outperform genre-selection in most cases, it can be helpful when in-genre data is not as linguistically diverse. For the targets SA-UFAL and MYV-JR (\textit{fiction}) both methods outperform genre-based selection by around 2\% LAS.

A clear example of insufficient in-genre data is the QHE-HIENCS target. It represents a highly-specialized variation of the \textit{social} genre, specifically Twitter data. Although the genre-based selection methods correctly identify and cluster the Italian Twitter data from IT-PoSTWITA \citep{sanguinetti2018} and IT-TWITTIRO \citep{cignarella2019}, there is a lack of such in-genre data from other languages,\footnote{More non-official Twitter-based treebanks in UD style exist \citep{sanguinetti-etal-2020-treebanking} which were left out of this study as they are not part of UD and contain annotation divergences.} leading these parsers to overfit on Italian specifically. This once again highlights the difficulty of selecting proxy training data which covers all desired characteristics --- even from a dataset as diverse as UD.

In general, the genre-driven methods make fairly similar selections given their shared baseline pool of treebanks containing the target genre in-mixture (see Appendix \ref{sec:selection-analysis}). Since using all of these data however results in the worst overall performance (\textsc{Meta}) while \textsc{Boot}, \textsc{GMM} and \textsc{LDA} perform best, the targeted selection of relevant subsets within the larger \textsc{Meta} pool appears to be key. Frequently, large treebanks such as Polish-LFG \citep{pl-lfg} with 14k instances from \textit{fiction}, \textit{news}, \textit{nonfiction}, \textit{social} and \textit{spoken} are subsampled to a much smaller fraction (around 3k instances in this example). The fact that these proportions as well as the selected instances themselves are relatively consistent across same-genre targets corroborates that all our methods are picking up on similar, data-driven notions of genre.

\begin{figure}
    \centering
    \begin{subfigure}[m]{.215\textwidth}
        \pdftooltip{\includegraphics[width=\textwidth]{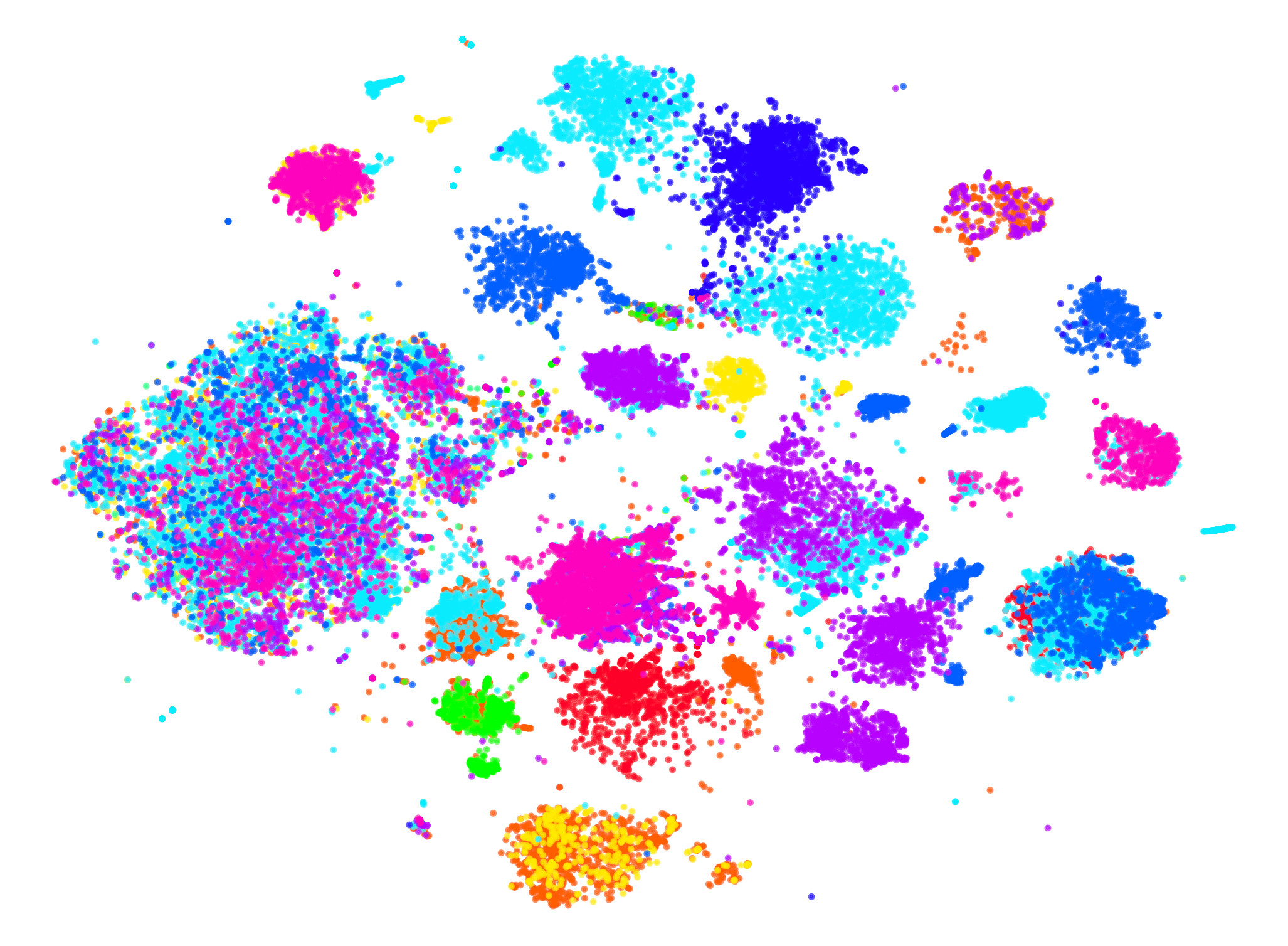}}{Screenreader Caption: tSNE plot of sentence embeddings from untuned mBERT. Two stand-alone news clusters top center, one stand-alone social cluster top center, remaining clusters mostly mixtures of two or more genres, center left has large cluster containing almost all genres in mixture.}
        \caption{mBERT}
        \label{fig:tsne-mbert-untuned-sds}
    \end{subfigure}
    \begin{subfigure}[m]{.215\textwidth}
        \pdftooltip{\includegraphics[width=\textwidth]{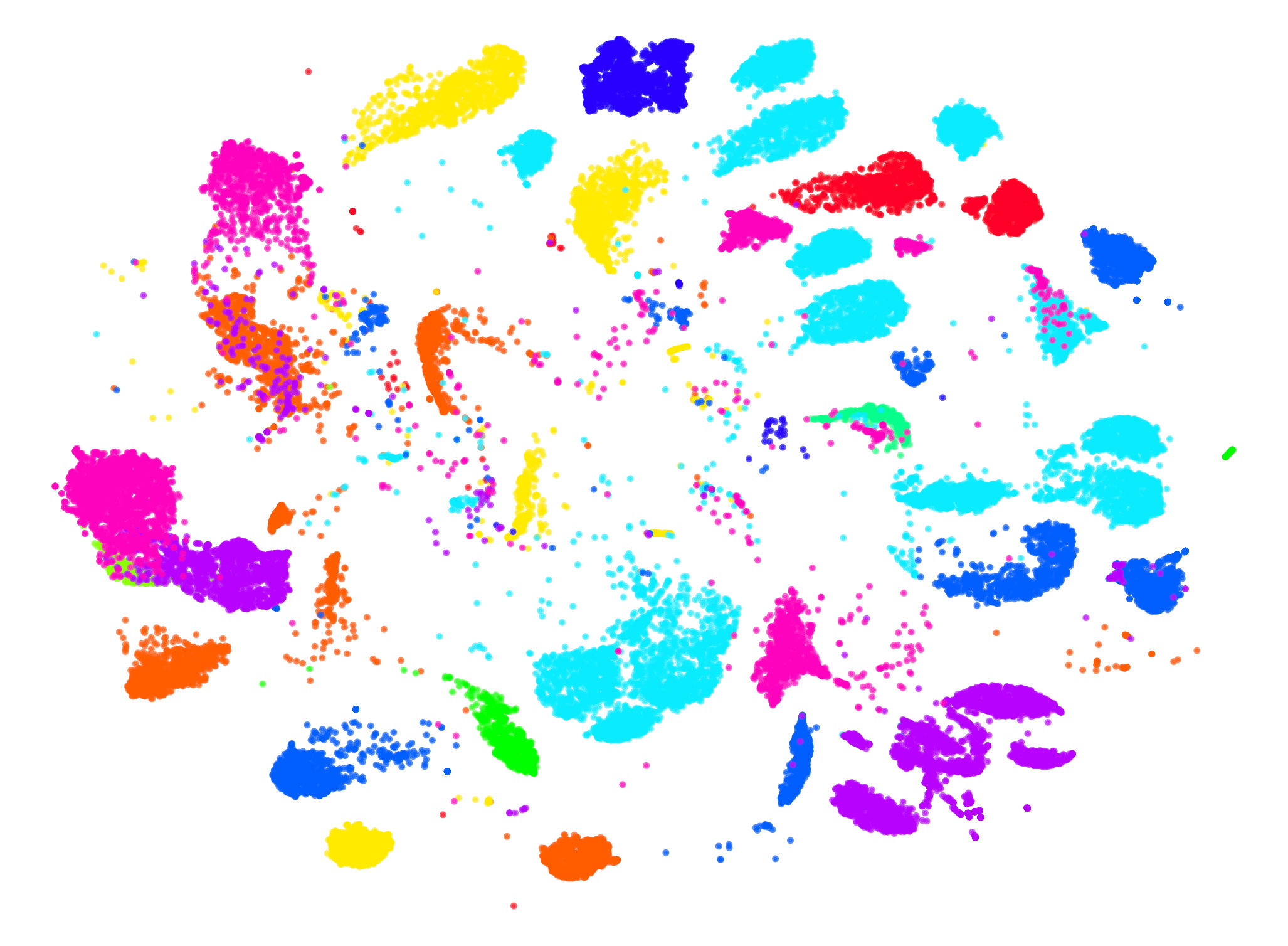}}{Screenreader Caption: tSNE plot of sentence embeddings from Boot-tuned mBERT. Almost no overlapping genre clusters. Each local group is genre-consistent.}
        \caption{\textsc{Boot}}
        \label{fig:tsne-mbert-boot-sds}
    \end{subfigure}\\[1em]
    \begin{subfigure}[m]{.45\textwidth}
        \centering
        \pdftooltip{\includegraphics[width=.9\textwidth]{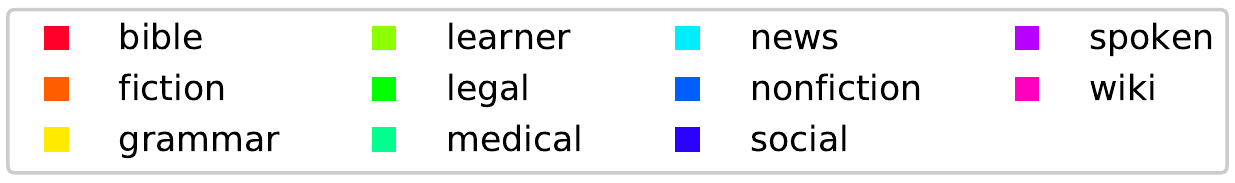}}{Screenreader Caption: Legend listing the genres bible, fiction, grammar, learner, legal, medical, news, nonfiction, social, spoken, wiki.}
        \vspace{.7em}
    \end{subfigure}
     \caption{\textbf{UD Genres in Embedding Space} of (a) untuned mBERT and (b) genre-tuned \boot{}. Sentences from single-genre treebanks (up to 1k each) colored by genre, plotted using tSNE \citep{maaten2008}. Tuning using genre as weak supervision clearly amplifies genre information.}
    \label{fig:tsne-mbert-sds}
\end{figure}

Figure \ref{fig:tsne-mbert-sds} further visualizes the presence of latent genre using t-SNE plots of up to 1k randomly sampled sentence embeddings from each of UD's single-genre treebanks. In their untuned state (Figure \ref{fig:tsne-mbert-untuned-sds}), some local genre clusters do manifest. However, these mainly correspond to specialized treebanks such as the aforementioned Italian Twitter treebanks (\textit{social}). Most other genres occur in language-level mixtures or in a large overall ``blob'' on the left. By amplifying genre explicitly using the \boot{} procedure, each individual genre is much more clearly segmented (Figure \ref{fig:tsne-mbert-boot-sds}).

In conclusion, the presence of similar performance patterns across all our proposed genre-driven methods --- while having separate approaches to treebank segmentation (weakly supervised tuning for \boot{}, treebank-internal embedding distances for \textsc{GMM}, n-grams for \textsc{LDA}) --- confirms our hypothesis that instance-level genre can be identified cross-lingually from contextualized representations and aids zero-shot parsing.

\section{Conclusions}\label{sec:conclusion}

In absence of in-language training data, we have explored UD-specified genre as an alternative signal for data selection. While prior work had indicated the presence of genre information in monolingual contextualized embeddings \citep{aharoni2020}, an analogous strategy using mBERT embeddings proved insufficient in the cross-lingual parsing setting (\textsc{Sen}), performing close to the random baseline (\textsc{Rand}). Relying on manual, treebank-level genre labels (\textsc{Meta}) proved even less performant, producing the lowest scores despite corresponding to a practitioner's typical first choice of selecting the largest number of training instances.

In order to enable finer-grained, instance-level data selection, we proposed three methods for combining latent genre information in the unsupervised contextualized representations with the treebank metadata: weakly supervised \boot{}, sentence embedding-based \textsc{GMM} and n-gram-based \textsc{LDA}. Despite their different approaches to treebank segmentation, each method significantly outperformed the purely embedding-based \textsc{Sent} as well as the metadata (\textsc{Meta}) and random baselines (\textsc{Rand}). Their similar performance patterns and selected data distributions further indicate that each method is identifying a shared, data-driven notion of genre.

For future work, it will be important to extend our proposed approaches beyond single-genre targets towards genre-mixtures and more treebanks overall. As the data selected by these methods is further limited by the number of treebanks in each respective genre, combining a larger set of selection signals will be equally crucial.

\section*{Acknowledgments}

We would like to thank the NLPnorth group for insightful discussions on this work --- in particular Elisa Bassignana, Maria Barrett and Mike Zhang. Thanks to Héctor Martínez Alonso for feedback on an early draft as well as ITU's High-performance Computing Cluster team. Finally, we thank the anonymous reviewers for their thorough feedback. This research is supported by the Independent Research Fund Denmark (DFF) grant 9063-00077B and an Amazon Faculty Research (ARA) Award.

\nocite{fontawesome}

\bibliography{anthology,references}
\bibliographystyle{acl_natbib}

\clearpage

%
%

\appendix\label{sec:appendix}

\section*{Appendix}

\section{Universal Dependencies Setup}\label{sec:ud-setup}
All experiments make use of Universal Dependencies v2.7 (\citealp{ud27}; UD). From the total set of 183 treebanks, we use all except for the following six (due to licensing restrictions): \textit{AR-NYUAD, EN-ESL, EN-GUMReddit, FR-FTB, JA-BCCWJ, GUN-Dooley}. In total 1.38 million sentences are used in our experiments.

\paragraph{Target Treebanks} As listed in the main paper, our target treebanks are \textit{Swedish Sign Language-SSLC} \citep{swl-sslc}, \textit{Sanskrit-UFAL} \citep{sa-ufal}, \textit{Komi Zyrian-Lattice} \citep{kpv-tbs}, \textit{Tamil-TTB} \citep{ta-ttb}, \textit{Galician-TreeGal} \citep{gl-treegal}, \textit{Cantonese-HK} \citep{yue-hk}, \textit{Chukchi-HSE} \citep{ckt-hse}, \textit{Faroese-OFT} \citep{fo-oft}, \textit{Telugu-MTG} \citep{te-mtg}, \textit{Erzya-JR} \citep{myv-jr}, \textit{Hindi-English-HIENCS} \citep{qhe-hiencs} and \textit{Turkish-German-SAGT} \citep{qtd-sagt}.

\paragraph{Development Data} For the initial tuning of LDA input features as well as the bootstrapping threshold, we used the only five treebanks with development data: \textit{SWL-SSLC, TA-TTB, TE-MTG, QHE-HIENCS, QTD-SAGT}.

For the early stopping of parser training, no such in-language validation data is used (to ensure a pure zero-shot setup). Instead, the data selected by each selection method is split in an 80\%/20\% fashion and is used as a proxy, out-of-language development split.

Similarly, the training of the bootstrapping classifier (\boot) uses only the non-target-language portion of UD (i.e.\ excluding all treebanks of the 12 target languages plus constituent languages for code-switched). For efficiency reasons, this data is further subsampled to 40k total instances. Both the training and validation (used for early stopping) of \boot{} are therefore similarly conducted without any target-language data.

\paragraph{Subsets} Since data selection is at the core of this research, the exact instance IDs of each subset are available in the supplementary code.

\section{Model and Training Details}\label{sec:training-details}

The following describes architecture and training details for all methods. When not further defined, default hyperparameters 
are used. Implementations are available in the supplementary code.

\paragraph{Infrastructure} Neural models are trained on an NVIDIA A100 GPU with 40 GB of VRAM. Since most of our experiments do not require MLM sentence embeddings to be updated, we compute them once and store them on disk to save GPU cycles.

\paragraph{Multilingual Language Model} The MLM used in this work is mBERT \citep{devlin2019} as implemented in the Transformers library \citep{wolf2020}\footnote{\verb@bert-base-multilingual-cased@}. Embeddings are of size $d_{\text{emb}}=768$ and the model itself has 178 million total parameters. To create sentence embeddings in the \textsc{Sent} and \textsc{GMM} methods, we use the mean-pooled WordPiece embeddings \citep{wu2016} of the final layer.

\paragraph{Clustering Methods} Both \textit{Gaussian Mixture Models} (\textsc{GMM}) and \textit{Latent Dirichlet Allocation} (\citealp{blei2003}; \textsc{LDA}) use implementations from scikit-learn v0.23 \citep{sklearn}. \textsc{LDA} uses bags of character 3--6-grams which occur in at least two and in at most 30\% of sentences. The n-gram sizes were initially tuned on target treebanks with available development sets (see Appendix \ref{sec:ud-setup}). We found character 1--5-grams to perform approximately 2.5 LAS worse and word unigrams to perform approximately 2 LAS worse than the final method. GMMs use the mBERT sentence embeddings directly as input. Both methods are CPU-bound and complete the clustering of all treebanks in UD in under 45 minutes.

\paragraph{Bootstrapping} (\textsc{Boot}) builds on the standard mBERT architecture as follows: mBERT $\rightarrow$ CLS $\rightarrow$ linear layer ($d_{\text{emb}} \times 18$) $\rightarrow$ softmax. The training has an epoch limit of 100 with early stopping after 3 iterations without improvements on the development set. No target-language data is used during this process. An alternate bootstrapping threshold of 0.9 was evaluated and found to perform approximately 1 LAS worse on the development subset (see Appendix \ref{sec:ud-setup}) than the final value of 0.99. Backpropagation is performed using AdamW \citep{loshchilov2017} with a learning rate of $10^{-7}$ on batches of size 16. The fine-tuning procedure requires GPU hardware which can host mBERT, corresponding to 10 GB of VRAM. Training on the subsampled 40k instance, non-target-language data takes approximately seven hours.

\paragraph{Dependency Parsers} Every parsing experiment in the main paper uses a biaffine attention parser \citep{dozat2017} implemented in the MaChAmp v0.2 framework \citep{machamp} using default hyperparameters. The sentence encoder is initialized with standard mBERT weights. The training duration is foremost dependent on input data quantity. For the largest corpus (\textsc{Meta} for TA-TTB with 524k instances) this corresponds to 55 hours. Our proposed methods create smaller, targeted training corpora (around 80k instances on average) such that a better performing parser can be trained in approximately 90 minutes on the same hardware.

\paragraph{Random Initializations} Each experiment is run thrice using the seeds 41, 42 and 43. This relates to the random subsampling of data as well as to model initialization (both parsers and selection).

\section{Additional Results}\label{sec:additional-results}

In addition to the labeled attachment scores (LAS) reported in the main paper, we list LAS standard deviation across random initializations in Table \ref{tab:parsing-results-stddev}, unlabeled attachment scores (UAS) in Table \ref{tab:parsing-results-uas} as well as the number of selected training instances per method in Table \ref{tab:parsing-results-size}.

\paragraph{Predictions} We additionally provide the instance-level predictions of each method and each random initialization as CoNLL-U files in the supplementary material in order ensure that future work can evaluate the statistical significance of performance differences.

\begin{table*}
\centering
\resizebox{\textwidth}{!}{
\begin{tabular}{lrrrrrrrrrrrrr}
\toprule
\textsc{Setup} & \multicolumn{1}{c}{\textsc{SWL} {\def\svgwidth{.65em}}} & \multicolumn{1}{c}{\textsc{SA} {\def\svgwidth{.6em}}} & \multicolumn{1}{c}{\textsc{KPV} {\def\svgwidth{.6em}}} & \multicolumn{1}{c}{\textsc{TA} {\def\svgwidth{.7em}}} & \multicolumn{1}{c}{\textsc{GL} {\def\svgwidth{.7em}}} & \multicolumn{1}{c}{\hspace{-.1em}\textsc{YUE} {\def\svgwidth{.65em}}} & \multicolumn{1}{c}{\hspace{-.1em}\textsc{CKT} {\def\svgwidth{.65em}}} & \multicolumn{1}{c}{\textsc{FO} {\def\svgwidth{.7em}}} & \multicolumn{1}{c}{\textsc{TE} {\def\svgwidth{.6em}}} & \multicolumn{1}{c}{\hspace{-.5em}\textsc{MYV} {\def\svgwidth{.6em}}} & \multicolumn{1}{c}{\hspace{-.1em}\textsc{QHE} {\def\svgwidth{.6em}}} & \multicolumn{1}{c}{\textsc{QTD} {\def\svgwidth{.65em}}} & \multicolumn{1}{|c}{\textsc{AVG}} \\
\midrule
\textsc{Target} & 87 & 3k & 132 & 400 & 600 & --- & --- & 1k & 1k & --- & 1k & 285 & \multicolumn{1}{|r}{839}\\
\midrule[.3pt]
\textsc{Rand} & 31k & 81k & 84k & 249k & 244k & 30k & 30k & 50k & 21k & 86k & 12k & 30k & \multicolumn{1}{|r}{79k} \\
\textsc{Sent} & 33k & 95k & 101k & 271k & 236k & 31k & 30k & 58k & 23k & 113k & 14k & 31k & \multicolumn{1}{|r}{86k} \\
\midrule[.3pt]
\midrule[.3pt]
\textsc{Meta} & 62k & 274k & 274k & 524k & 523k & 62k & 62k & 125k & 35k & 274k & 57k & 61k & \multicolumn{1}{|r}{194k} \\
\midrule[.3pt]
\textsc{Boot} & 29k & 59k & 59k & 256k & 254k & 28k & 28k & 35k & 21k & 58k & 7k & 29k & \multicolumn{1}{|r}{72k} \\
\midrule[.3pt]
\textsc{GMM} & 33k & 95k & 101k & 271k & 236k & 31k & 30k & 58k & 23k & 113k & 14k & 31k & \multicolumn{1}{|r}{86k} \\
\textsc{LDA} & 32k & 89k & 95k & 238k & 233k & 33k & 33k & 56k & 21k & 96k & 14k & 30k & \multicolumn{1}{|r}{81k} \\
\bottomrule
\end{tabular}
}
\caption{\label{tab:parsing-results-size} \textbf{Training Corpus Sizes} (number of selected instances) for zero-shot parsing experiments from target/proxy in-language treebanks (\textsc{Target}; where available), random sentence selection (\textsc{Rand}) and closest sentence selection (\textsc{Sent}), treebanks containing target genre (\textsc{Meta}), instances classified as target genre (\textsc{Boot}), closest cluster selection (\textsc{GMM} and \textsc{LDA}).}
\end{table*}

\begin{table*}
\centering
\resizebox{\textwidth}{!}{
\begin{tabular}{lrrrrrrrrrrrrr}
\toprule
\textsc{Setup} & \multicolumn{1}{c}{\textsc{SWL} {\def\svgwidth{.65em}}} & \multicolumn{1}{c}{\textsc{SA} {\def\svgwidth{.6em}}} & \multicolumn{1}{c}{\textsc{KPV} {\def\svgwidth{.6em}}} & \multicolumn{1}{c}{\textsc{TA} {\def\svgwidth{.7em}}} & \multicolumn{1}{c}{\textsc{GL} {\def\svgwidth{.7em}}} & \multicolumn{1}{c}{\hspace{-.1em}\textsc{YUE} {\def\svgwidth{.65em}}} & \multicolumn{1}{c}{\hspace{-.1em}\textsc{CKT} {\def\svgwidth{.65em}}} & \multicolumn{1}{c}{\textsc{FO} {\def\svgwidth{.7em}}} & \multicolumn{1}{c}{\textsc{TE} {\def\svgwidth{.6em}}} & \multicolumn{1}{c}{\hspace{-.5em}\textsc{MYV} {\def\svgwidth{.6em}}} & \multicolumn{1}{c}{\hspace{-.1em}\textsc{QHE} {\def\svgwidth{.6em}}} & \multicolumn{1}{c}{\textsc{QTD} {\def\svgwidth{.65em}}} & \multicolumn{1}{|c}{\textsc{AVG}} \\
\midrule
\textsc{Target} & 40.66 & 38.74 & 26.70 & 75.83 & 85.51 & --- & --- & 58.78 & 91.26 & --- & 73.62 & 66.75 & \multicolumn{1}{|r}{61.98}\\
\midrule[.3pt]
\textsc{Rand} & 22.81 & \textbf{47.06} & 25.97 & 72.14 & \textbf{84.68} & 49.70 & 29.39 & 71.66 & 83.73 & 36.88 & \textbf{40.63} & 58.97 & \multicolumn{1}{|r}{51.97} \\
\textsc{Sent} & 24.47 & 44.98 & 31.69 & 71.28 & 84.63 & 51.11 & 31.95 & 71.92 & 83.03 & \textbf{41.73} & 40.19 & 58.85 & \multicolumn{1}{|r}{52.99} \\
\midrule[.3pt]
\midrule[.3pt]
\textsc{Meta} & 24.94 & 44.62 & 25.77 & 72.26 & 84.26 & 47.91 & 22.66 & 70.54 & 82.06 & 36.67 & 19.83 & 57.93 & \multicolumn{1}{|r}{49.12} \\
\midrule[.3pt]
\textsc{Boot} & 24.83 & 42.00 & 39.40 & 73.38 & 84.19 & \textbf{60.72} & 35.42 & 75.21 & 84.05 & 39.03 & 27.59 & 57.15 & \multicolumn{1}{|r}{53.58} \\
\midrule[.3pt]
\textsc{GMM} & 25.18 & 44.19 & 37.77 & \textbf{74.33} & 84.55 & 60.61 & \textbf{37.53} & 77.00 & 82.89 & 38.09 & 26.65 & \textbf{59.52} & \multicolumn{1}{|r}{54.02} \\
\textsc{LDA} & \textbf{27.42} & 44.84 & \textbf{40.33} & 72.93 & 84.27 & 60.06 & 35.68 & \textbf{77.23} & \textbf{84.70} & 38.78 & 27.61 & 58.46 & \multicolumn{1}{|r}{\textbf{54.36}} \\
\bottomrule
\end{tabular}
}
\caption{\label{tab:parsing-results-uas} \textbf{Unlabeled Attachment Scores} for zero-shot parsing experiments on test splits of target treebanks using training data from from target/proxy in-language treebanks (\textsc{Target}; where available), random sentence selection (\textsc{Rand}) and closest sentence selection (\textsc{Sent}), treebanks containing target genre (\textsc{Meta}), instances classified as target genre (\textsc{Boot}), closest cluster selection (\textsc{GMM} and \textsc{LDA}).}
\end{table*}

\begin{table*}
\centering
\resizebox{\textwidth}{!}{
\begin{tabular}{lrrrrrrrrrrrrr}
\toprule
\textsc{Setup} & \multicolumn{1}{c}{\textsc{SWL} {\def\svgwidth{.65em}}} & \multicolumn{1}{c}{\textsc{SA} {\def\svgwidth{.6em}}} & \multicolumn{1}{c}{\textsc{KPV} {\def\svgwidth{.6em}}} & \multicolumn{1}{c}{\textsc{TA} {\def\svgwidth{.7em}}} & \multicolumn{1}{c}{\textsc{GL} {\def\svgwidth{.7em}}} & \multicolumn{1}{c}{\hspace{-.1em}\textsc{YUE} {\def\svgwidth{.65em}}} & \multicolumn{1}{c}{\hspace{-.1em}\textsc{CKT} {\def\svgwidth{.65em}}} & \multicolumn{1}{c}{\textsc{FO} {\def\svgwidth{.7em}}} & \multicolumn{1}{c}{\textsc{TE} {\def\svgwidth{.6em}}} & \multicolumn{1}{c}{\hspace{-.5em}\textsc{MYV} {\def\svgwidth{.6em}}} & \multicolumn{1}{c}{\hspace{-.1em}\textsc{QHE} {\def\svgwidth{.6em}}} & \multicolumn{1}{c}{\textsc{QTD} {\def\svgwidth{.65em}}} & \multicolumn{1}{|c}{\textsc{AVG}} \\
\midrule
\textsc{Target} & 0.71 & 0.54 & 0.77 & 1.16 & 0.24 & --- & --- & 1.32 & 0.97 & --- & 0.26 & 1.10 & \multicolumn{1}{|r}{0.79}\\
\midrule[.3pt]
\textsc{Rand} & 1.60 & 0.46 & 0.16 & 0.72 & 0.09 & 1.33 & 0.89 & 1.02 & 0.64 & 1.09 & 0.55 & 0.55 & \multicolumn{1}{|r}{0.76} \\
\textsc{Sent} & 2.13 & 2.00 & 0.58 & 1.76 & 0.18 & 0.67 & 0.27 & 0.63 & 0.92 & 0.37 & 0.37 & 0.91 & \multicolumn{1}{|r}{0.90} \\
\midrule[.3pt]
\midrule[.3pt]
\textsc{Meta} & 0.90 & 0.75 & 0.73 & 1.24 & 0.27 & 0.41 & 1.19 & 0.82 & 0.42 & 0.44 & 0.44 & 0.73 & \multicolumn{1}{|r}{0.73} \\
\midrule[.3pt]
\textsc{Boot} & 0.54 & 0.85 & 0.55 & 1.07 & 0.27 & 0.14 & 0.51 & 0.92 & 0.42 & 0.28 & 1.08 & 0.43 & \multicolumn{1}{|r}{0.59} \\
\midrule[.3pt]
\textsc{GMM} & 1.14 & 1.02 & 0.75 & 1.00 & 0.18 & 0.28 & 0.80 & 1.30 & 1.35 & 1.28 & 0.63 & 0.47 & \multicolumn{1}{|r}{0.85} \\
\textsc{LDA} & 0.74 & 2.29 & 0.23 & 1.96 & 0.14 & 0.65 & 1.32 & 0.41 & 0.44 & 0.81 & 1.23 & 0.25 & \multicolumn{1}{|r}{0.87} \\
\bottomrule
\end{tabular}
}
\caption{\label{tab:parsing-results-stddev} \textbf{Standard Deviations of LAS} for zero-shot parsing experiments on test splits of target treebanks using training data from from target/proxy in-language treebanks (\textsc{Target}; where available), random sentence selection (\textsc{Rand}) and closest sentence selection (\textsc{Sent}), treebanks containing target genre (\textsc{Meta}), instances classified as target genre (\textsc{Boot}), closest cluster selection (\textsc{GMM} and \textsc{LDA}).}
\end{table*}

\section{Data Selection Analysis}\label{sec:selection-analysis}

\begin{figure*}
    \centering
    \includegraphics[height=235mm]{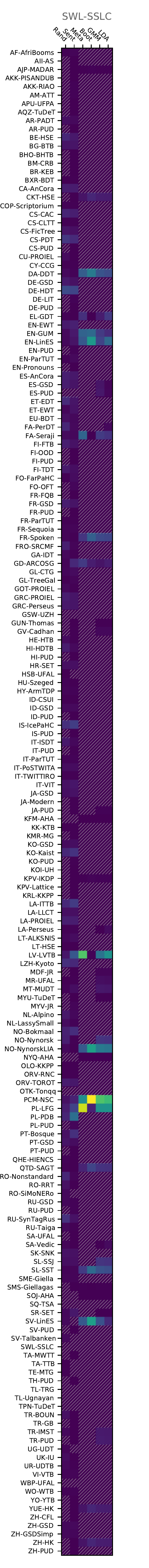}
    \hspace{-1.75em}
    \includegraphics[height=235mm]{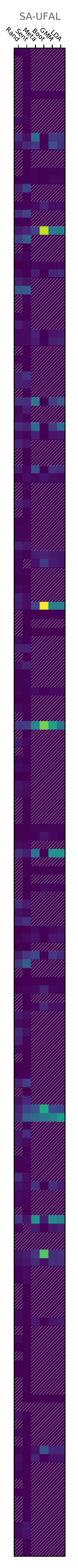}
    \hspace{-.75em}
    \includegraphics[height=235mm]{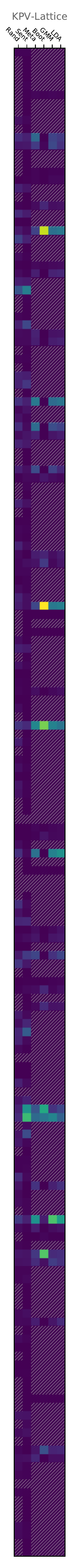}
    \hspace{-.75em}
    \includegraphics[height=235mm]{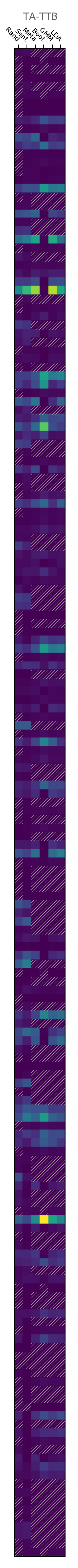}
    \hspace{-.75em}
    \includegraphics[height=235mm]{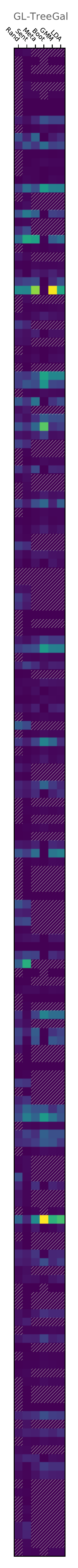}
    \hspace{-.75em}
    \includegraphics[height=235mm]{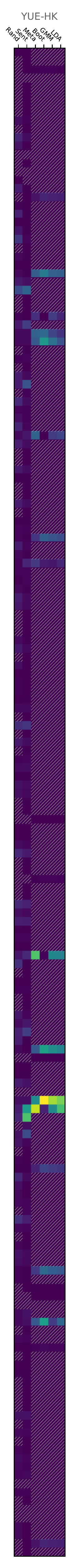}
    \hspace{-.75em}
    \includegraphics[height=235mm]{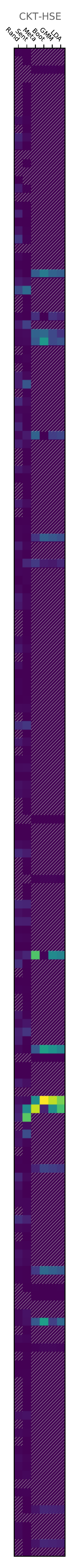}
    \hspace{-.75em}
    \includegraphics[height=235mm]{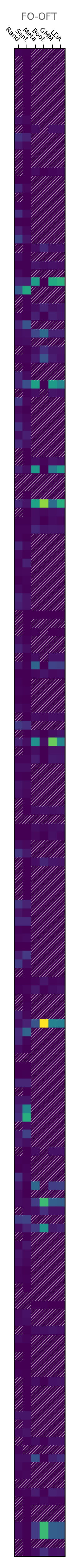}
    \hspace{-.75em}
    \includegraphics[height=235mm]{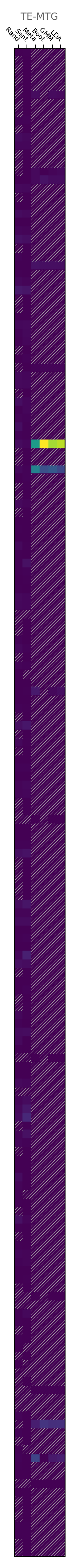}
    \hspace{-.75em}
    \includegraphics[height=235mm]{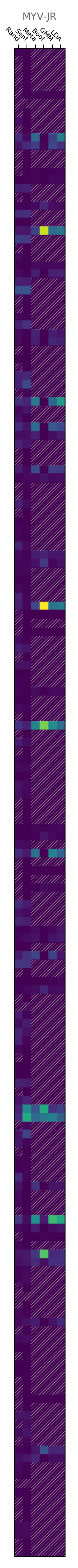}
    \hspace{-.75em}
    \includegraphics[height=235mm]{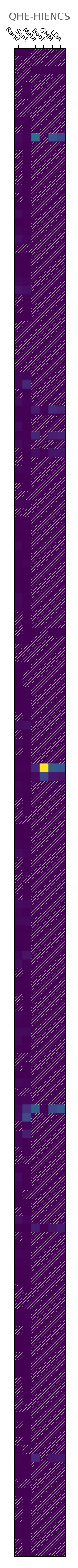}
    \hspace{-.75em}
    \includegraphics[height=235mm]{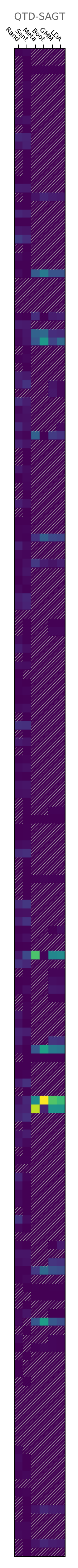}
    \caption{\textbf{Selection Proportions} per target treebank and data selection method across all of UD. Zero instances were selected from shaded regions.}
    \label{fig:selections}
\end{figure*}

Figure \ref{fig:selections} displays the distribution of selected instances across all treebanks of UD per target treebank and method. Proportions are normalized to $[0,1]$ for each method (i.e.\ across each column). Due to the large number of cells, we recommend viewing this figure digitally.

\end{document}